\documentclass[preprint,12pt]{elsarticle}

\usepackage{amssymb}
\usepackage{amsmath}
\usepackage{amsthm}

\usepackage{xcolor}
\usepackage{multirow}
\usepackage{multicol}
\usepackage{array}
\usepackage{hyperref}
\usepackage{graphicx}
\DeclareGraphicsExtensions{.png,PNG,.pdf,.PDF,.eps}
\usepackage{booktabs}
\usepackage{csquotes}
\usepackage{caption}
\usepackage{subcaption}
\usepackage{makecell}
\usepackage{tabularx}
\usepackage{adjustbox}
\usepackage{soul}
\usepackage{enumitem}
\usepackage{siunitx}
\usepackage{dsfont}

\usepackage{lineno}

\journal{Automation in Construction}

\begin{document}

\begin{frontmatter}



\title{Learning to Capture Rocks using an Excavator: A Reinforcement Learning Approach with Guiding Reward Formulation}





\author[label1]{Amirmasoud Molaei\corref{mycorrespondingauthor}}
\cortext[mycorrespondingauthor]{Corresponding author}\ead{amirmasoud.molaei@tuni.fi}
\author[label1]{Mohammad Heravi}
\author[label1]{Reza Ghabcheloo}
\affiliation[label1]{organization={Faculty of Engineering and Natural Sciences, Tampere University},
            city={Tampere},
            postcode={33720},
            country={Finland}} 

\begin{abstract}

Rock capturing with standard excavator buckets is a challenging task typically requiring the expertise of skilled operators. Unlike soil digging, it involves manipulating large, irregular rocks in unstructured environments where complex contact interactions with granular material make model-based control impractical. Existing autonomous excavation methods focus mainly on continuous media or rely on specialized grippers, limiting their applicability to real-world construction sites. This paper introduces a fully data-driven control framework for rock capturing that eliminates the need for explicit modeling of rock or soil properties. A model-free reinforcement learning agent is trained in the AGX Dynamics\textsuperscript{\textregistered} simulator using the Proximal Policy Optimization (PPO) algorithm and a guiding reward formulation. The learned policy outputs joint velocity commands directly to the boom, arm, and bucket of a CAT\textsuperscript{\textregistered}365 excavator model. Robustness is enhanced through extensive domain randomization of rock geometry, density, and mass, as well as the initial configurations of the bucket, rock, and goal position. To the best of our knowledge, this is the first study to develop and evaluate an RL-based controller for the rock capturing task. Experimental results show that the policy generalizes well to unseen rocks and varying soil conditions, achieving high success rates comparable to those of human participants while maintaining machine stability. These findings demonstrate the feasibility of learning-based excavation strategies for discrete object manipulation without requiring specialized hardware or detailed material models.


\end{abstract}




\end{frontmatter}


\section{Introduction}\label{sec:intro}
Autonomous excavation holds a great promise in addressing increasing demands of the mining and construction industries, two of the largest and most essential sectors worldwide. The excavator is one of the most widely used and versatile heavy-duty mobile machines (HDMMs), which is typically operated through a hydraulic system. Excavators are utilized for a wide range of earth-moving tasks, including digging, trenching, grading, and in particular material handling. Despite their versatility, traditional manual operation of excavators can result in low efficiency, increased physical strain on operators, and exposure to hazardous environments like open-pit mines. These challenges underscore the need for automation to enhance safety and productivity. An excavator is primarily composed of three major components, the traveling body, swing body, and the front digging manipulator. The digging manipulator, includes three main parts, boom, arm, and bucket, which are actuated by hydraulic cylinders. Additionally, joints connect the swing body, boom, arm, and bucket, allowing for flexible and precise motion~\cite{Autonomous2020Sotiropoulos,egli2024reinforcement,Ishmatuka2023Autonomous,Molaei2023AutomaticLoading}.

Bulk material handling is effectively performed by HDMMs, particularly excavators. In large-scale open-pit mining, where ledges are blasted, or on construction sites where rocks are blocked the working area or buried in soil, excavators often face the challenge of picking individual large rocks mixed with finer gravel and soil. Collecting these large rocks and transporting them to dump trucks or other destinations requires high degree of expertise, typically only experienced operators have~\cite{Werner2024Dynamic,Autonomous2020Sotiropoulos,Egli2022General,Liangjun2021autonomous}.

\begin{figure}[htbp]
\begingroup
\captionsetup[subfigure]{skip=0pt, margin=0pt}
\centering
\begin{subfigure}[t]{0.4\linewidth}
    \centering
    \includegraphics[width=\linewidth]{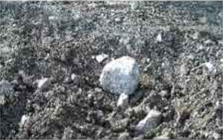}
    \caption{}
    \label{fig:schematicRockRemoval01}
\end{subfigure}
\begin{subfigure}[t]{0.4\linewidth}
    \centering
    \includegraphics[width=\linewidth]{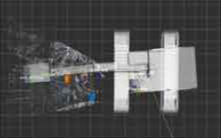}
    \caption{}
    \label{fig:schematicRockRemoval02}
\end{subfigure}
\vspace{0pt} 
\begin{subfigure}[t]{0.4\linewidth}
    \centering
    \includegraphics[width=\linewidth]{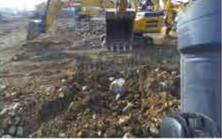}
    \caption{}
    \label{fig:schematicRockRemoval03}
\end{subfigure}
\begin{subfigure}[t]{0.4\linewidth}
    \centering
    \includegraphics[width=\linewidth]{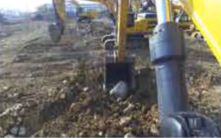}
    \caption{}
    \label{fig:schematicRockRemoval04}
\end{subfigure}
\caption{Rock capturing task using an excavator~\cite{Liangjun2021autonomous}.}
\label{fig:schematicRockRemoval}
\endgroup
\end{figure}

Autonomous rock capturing presents unique challenges due to the different behavior of large rocks compared to homogeneous materials. While homogeneous materials can often be excavated in a predefined scooping motion, rocks must be individually and precisely captured using the bucket, requiring a different control strategy. Figure~\ref{fig:schematicRockRemoval} illustrate the rock capturing task. A major difficulty lies in the complex and poorly understood interactions between the bucket and surrounding materials, especially when both soil and rock are present~\cite{Huang1993TOWARDAA}. Modeling these interactions using physical models, such as terra-mechanics models~\cite{Blouin2001Review}, is highly challenging due to unknown and variable factors such as rock mass, geometry, friction, and material properties. Moreover, the terrain may be uneven, and critical parameters related to the material are often unavailable or difficult to estimate accurately~\cite{Robotic1993Xiaodong,Automatic2014McKinnon}. Even with accurate knowledge of these parameters, designing controllers in the traditional way remains highly challenging. Therefore, machine learning algorithms, such as Reinforcement Learning (RL), can be a promising solution to this complex task.

Identifying the similarities between robotic manipulation and rock excavation offers valuable insight into the rock capturing task, indicating that techniques developed in industrial manipulator literature can be applied to this task. Capturing a rock using a bucket has similarities to a non-prehensile manipulation task in the presence of granular material~\cite{Autonomous2020Sotiropoulos}. In this task, the bucket interacts with the environment to guide the rock along a desired trajectory. Research in non-prehensile manipulation has made significant progress in recent years, with several studies extensively investigating RL approaches for acquiring such skills~\cite{Del2023Nonprehensile,Zhu2022Excavation,Wang2024Multi}.



RL techniques are based on the concept of an agent, a program that interacts with the environment by performing actions aimed at maximizing a predefined cumulative reward. Through continuous exploration, the agent gathers information from observations and receives feedback in the form of rewards, which guide its learning process. Positive rewards reinforce desirable behavior, while negative rewards discourage incorrect actions. Moreover, RL agents can be retrained for new patterns and enabling further improvements~\cite{Kurinov2020Automated}. Training a control policy using RL demands a large number of interaction samples, which are often impractical to collect using physical robots. To address this, physics-based simulators, such as Bullet Physics~\cite{Coumans2015Bullet} and MuJoCo~\cite{Todorov2012mujoco}, are commonly used as training environments for RL agents. However, these simulators generally face challenges in accurately representing granular materials~\cite{Collins2021Review}. It can cause a significant challenge regarding training an efficient control policy for the rock capturing task, since the interactions of granular materials with the bucket and rock are highly essential.

In this paper, a learning-based approach is proposed to automatically capture large rocks using an excavator bucket, without relying on any explicit knowledge of the rock geometry or mass or surrounding material properties. A control policy is trained using model-free RL within the high-fidelity AGX Dynamics\textsuperscript{\textregistered} simulation environment. Prior research has shown that AGX Dynamics\textsuperscript{\textregistered} offers the realism and performance necessary for simulating complex earth-moving tasks involving granular media and rigid body interactions~\cite{Servin2021Multiscale,tera2024Aluckal,Aoshima2024Sim2Real}. In our framework, a Proximal Policy Optimization (PPO) algorithm is employed alongside guiding reward formulation that encourages successful rock manipulation while maintaining machine stability. To enhance the robustness and generalizability of the learned policy, key parameters such as rock geometry, density, mass, initial positions of the rock and bucket, and goal position are extensively randomized during training. The trained policy is evaluated under various conditions, including scenarios with previously unseen rock shapes and different material properties, and demonstrates generalization capabilities. Results indicate that the controller can adapt its strategy effectively to novel situations and achieves faster task completion compared to human participants.

The rest of this paper is structured as follows. Section~\ref{sec:LitRev} provides a review of the literature. Section~\ref{sec:Methodolgy} outlines the proposed approach, and the results are presented in Section~\ref{sec:results}. Section~\ref{sec:Discussion} discusses the characteristics of the proposed method, and the conclusions are provided in Section~\ref{sec:Conclusion}.

\section{Literature Review} \label{sec:LitRev}

Over the past several years, many studies have explored the automation of excavators, with particular attention given to the bucket-filling task. Nevertheless, autonomous excavators still fall short of matching the performance and flexibility of skilled human operators, which continues to limit their widespread adoption in industries~\cite{Johnson2023AutonomousConstruction,egli2024reinforcement}. In addition, the rock capturing task, a  challenging and safety-critical operation, has largely been overlooked in the literature~\cite{Autonomous2020Sotiropoulos}.

A straightforward approach to automate the bucket-filling task involves designing a predefined bucket trajectory through the soil~\cite{Modeling1996Koivo,Yoshida2021Practical,Shao2008Automatic}. While this method performs well under uniform soil conditions, it tends to fail when those conditions vary~\cite{egli2024reinforcement}.
The excavation process can be divided into several steps, such as penetration, dragging, scooping, and lifting. To minimize manual adjustments, in~\cite{Data2020Sandzimier}, a controller is proposed that automatically transitions from dragging to scooping for more accurate bucket filling. However, essential parameters, especially the excavation depth of the trajectory, still require manual tuning for a specific soil type. 
To address it, in~\cite{Sotiropoulos2019Model}, a method  is suggested to adjust the bucket height during the dragging phase to apply maximum power to the soil, improving efficiency and preventing stalling. Despite this, other parts of the trajectory still require manual calibration. 
In~\cite{Sotiropoulos2022Dynamic}, a Model Predictive Control (MPC) approach is introduced to follow the desired shape. This approach utilizes a soil-bucket model based on the Koopman theory, trained on data collected from excavation experiments involving a single soil type. However, this controller requires precise force control, which demands costly modifications to the hydraulic system~\cite{Hutter2015Towards}. Furthermore, the extension of this method to handle various soil types remains an open question for future research. 
In~\cite{Jud2017Planning,Jud2019Autonomous}, by defining a bucket-force trajectory for each excavation phase, a soil-adaptive algorithm is obtained across different soils. It could accurately excavate a target shape by switching to position control once the bucket reaches the desired height. Accurate force control is obtained by replacing the standard main-stage valves with high-performance servo valves. 
In~\cite{reginald2021integrative}, an impedance control is employed to track the desired bucket force to prevent stalling while following a predefined bucket trajectory. However, this results in incomplete bucket filling when the soil is harder than expected. 
This issue is addressed in~\cite{Maeda2015Combined} by proposing an iterative learning controller with a disturbance observer. The method assumes similar soil responses, which does not always hold true. 
In~\cite{Park2017Online}, an Echo-State Network (ESN) is used to learn an inverse model of an excavator for trajectory tracking in repetitive digging tasks. The ESN is initially pretrained with a conventional Proportional-Derivative (PD) controller and then updated online during operation to improve performance under varying conditions. While the controller is able to adapt over time, it requires multiple digging cycles and assumes consistent soil conditions.

Rule-based approaches are another strategy to prevent stalling, typically by implementing predefined corrective actions when interaction forces become excessive~\cite{park2017implementation} or when tracking errors grow too large~\cite{Dunbabin2006Autonomous}. Some studies have investigated this concept by creating libraries of motion primitives, which are then selected or combined using rule-based logic to perform autonomous excavation tasks~\cite{bradley1998development,groll2019autonomous,Quang2002Robotic,Schmidt2010Simulation,Xiaobo1996Experimental}. Although such methods have been successfully implemented on real machines, developing the motion primitives demands significant engineering work and deep domain expertise, particularly for handling more complex operations~\cite{egli2024reinforcement}.

To move beyond manually designed bucket trajectories, Trajectory Optimization (TO) methods have been widely explored for autonomous excavation. Generally, kinematic-based approaches optimize objective functions, including the volume of scooped soil, time, and motion smoothness~\cite{Yang2021Optimization,Yunyue2021Time}. Although these methods can consider both the current and target terrain elevations and are suitable for real-time applications, they fall short in ensuring trajectory feasibility since they neglect soil reaction forces. 
To address it, a dynamics-aware MPC framework with a disturbance observer is introduced in~\cite{Lee2021Real} to follow a kinematically optimized trajectory and demonstrated in simulation. However, when disturbances arise, such as encountering soil that is harder than expected, the system diverges from the intended path, leading to incomplete bucket filling, as the trajectory is not updated in real time. Other TO methods incorporate soil properties by embedding soil models into the optimization process, which significantly increases computational complexity, making real-time execution impractical~\cite{KIM2013Dynamically,Time2020Yang,Yoshida2013DiggingTrajectory,Zhihong2019Task}.
To reduce the computational time in deployment, in~\cite{Time2020Yang}, an approach is suggested distilling TO results into a neural network, which enables fast inference within milliseconds.
A common limitation of TO approaches, that rely on soil models, is the need for prior knowledge of soil parameters. These parameters can be obtained through time-consuming geological site inspection~\cite{Zhihong2019Task}, or by optimizing model parameters to minimize the error between predicted and actual soil forces during either manual operation~\cite{KIM2013Dynamically}, or laboratory experiments~\cite{Hybrid2009Althoefer}. 
In~\cite{Zhao2020IEEE}, a supervised learning approach is used to estimate soil parameters directly from measured resistance data while excavating known soil types. However, the accuracy of these methods depends heavily on the quantity and quality of the training data, which are time-intensive to gather. Additionally, it is still uncertain whether these models can generalize across different machines.

Rather than relying on TO to generate excavation trajectory, another line of research utilizes expert demonstrations. These demonstrations are either reparameterized~\cite{Zhao2022Spline} or optimized using stochastic methods~\cite{Guo2022Imitation} to achieve kinematic goals such as speed or smoothness. However, since these approaches neglect soil reaction forces, they risk causing the excavator to stall. One way to address it is by imposing a force threshold and restricting the excavation area that can result in inefficient excavation~\cite{Huh2023Deep}.
In~\cite{Son2020Expert}, a demonstration-based method is introduced to learn dynamic motion primitive parameters for replicating human excavation trajectories, featuring online adaptation to adjust excavation depth and prevent excessive forces or stalling~. However, because the trajectory endpoint is fixed, the approach frequently leads to insufficient bucket filling, especially in harder soils where interaction forces are greater~\cite{Egli2022Soil}.
The proposed methods in~\cite{schenck2017learning,Lu2021Excavation} involve training visual prediction models of the excavation scene using Convolutional Neural Network (CNN) and data collected from real machines and simulations. The model is then used in sampling-based optimization to determine suitable actions for reaching a desired state of scene without accounting for interaction forces.
In~\cite{Fukui2017Imitation}, a method is proposed using imitation learning to automate excavation by classifying and adapting expert-demonstrated motions. However, the approach depends on a large database of excavation, which is challenging to gather in real-world.
In~\cite{Tahara2022Disturbance}, an imitation learning approach is developed capable of utilizing non-optimal demonstrations. The method is applied to an excavator model operating in granular soil. A persistent challenge across these techniques is their heavy reliance on the quantity, quality, and diversity of demonstration data, raising questions about their transferability and effectiveness across different excavators~\cite{egli2024reinforcement}.

To eliminate the need for costly and time-consuming data collection on real machines, RL in simulation offers a promising solution. 
In~\cite{Zhu2022Excavation}, an offline RL algorithm is applied on a Franka robot, introducing a cost function to discourage large reaction forces in the optimization of expert demonstrations. While this method generally results in trajectories requiring less torque, it does not guarantee the feasibility of the trajectory. 
Similarly, in~\cite{jin2023learningexcavationrigidobjects} an offline RL is employed to minimize forces while maximizing bucket depth in the penetration phase. 
In~\cite{Lu2022Excavation}, an RL policy is trained in simulation for excavating rigid objects using a Franka Panda arm, relying on visual input from the excavation scene. However, due to the absence of soil interaction forces, the policy frequently failed during deployment because of discrepancies between simulation and real-world conditions. 
In~\cite{Samtani2023Learning}, an RL controller is trained to push against a rock and activate an excavator-mounted hammer tool for rock breaking. The sim-to-real gap is addressed by incorporating actuation delays in the simulation and applying filtering to sensor measurements during deployment. 
In~\cite{Zhu2017EnergyRL}, RL is also applied to real-time energy management in a hybrid excavator, where it regulates energy flow and enhance efficiency.
In~\cite{Benjamin2018Learning}, the PPO algorithm is used to train a neural network agent to perform a surface leveling task. The simulation is performed using the Dynasty simulation engine (proprietary software by Caterpillar Inc.), while the transfer to the real world is left for future work.
In~\cite{Kurinov2020Automated}, RL is used to train a neural network controller using multibody excavator model. 
In~\cite{Egli2022General}, RL is used to precisely control an excavator manipulator for the grading task. To reduce the sim-to-real gap and reach competitive accuracy, the simulation accuracy is enhanced with an actuator model trained using real data. 
In~\cite{Egli2022Soil}, a simple analytical soil model based on the Fundamental Equation of Earth-moving (FEE)~\cite{Reece1964Fundamental} is employed to train an RL controller.

Given the substantial sample requirements for RL training, these controllers are typically trained in simulation before being transferred to physical hardware. Moreover, testing with HDMMs, such as excavators, is both costly and potentially hazardous, making simulation-based experiments a crucial preliminary step. Popular examples of these simulators include Bullet Physics, MuJoCo, Open Dynamics Engine (ODE), NVIDIA PhysX, and Havok. While simulators like PhysX, Havok, and ODE often lack adequate precision for high-accuracy applications~\cite{Erez2015Simulation}, Bullet Physics and MuJoCo also face limitations when modeling interactions involving soft or granular materials~\cite{Collins2021Review}. Due to the difficulty of achieving sufficiently accurate simulations, successful sim-to-real transfer for conventional construction machinery remains rare and has not yet been demonstrated in studies such as~\cite{Kurinov2020Automated,Matsumoto2020Simulation,Osa2022Deep}. One potential solution to these challenges is the use of high-fidelity physics-engine simulators. Tools like AGX Dynamics\textsuperscript{\textregistered} can simulate the excavation process with high accuracy. In~\cite{Servin2021MultiscaleTerrain}, a multi-scale terrain simulation method based on AGX Dynamics\textsuperscript{\textregistered} is shown to replicate excavation forces and soil displacement with an error margin of 10-25\%, while still achieving real-time performance. 
In~\cite{tera2024Aluckal}, a simulation platform called TERA, built with Unity3D and AGX Dynamics\textsuperscript{\textregistered}, is developed to accurately model excavator–terrain interaction and enable scalable simulations. 
In~\cite{Aoshima2024Sim2Real}, a simulation-to-reality discrepancy of roughly 10\% has been reported in bucket-filling tasks. Also, in~\cite{Aoshima2025High}, AGX Dynamics\textsuperscript{\textregistered}-based simulations integrated with world models is used to optimize sequential loading operations. Collectively, these results have demonstrated that AGX Dynamics\textsuperscript{\textregistered} delivers the precision required for simulating excavation processes.

Capturing a rock using a bucket of an excavator is considered a non-prehensile manipulation task in the presence of granular material within the robotics literature~\cite{Matthew1999Progress}. In this task, the bucket, which is functioning as a non-prehensile end-effector, must move the rock by interacting with the surrounding environment. Actually, the interaction between the bucket and the rock can be transmitted indirectly through the surrounding material. In addition, the behavior of materials introduces considerable complexity and is inherently challenging to model. Moreover, the motion of the rock does not follow a flat ground plane due to the uneven and dynamically changing soil conditions~\cite{Autonomous2020Sotiropoulos}. Although the excavation of individual rocks using excavators is recognized as a challenging task in~\cite{Robotic1993Xiaodong,Huang1993TOWARDAA,Automatic2014McKinnon}, only two approaches have been introduced in~\cite{bradley1998development,Autonomous2020Sotiropoulos}.
In~\cite{bradley1998development}, a trial-and-error method is proposed, where the rock is removed by digging at progressively deeper depths. The approach relies on a set of if-then rules designed to mimic skilled operators and is calibrated on a one-fifth-scale LUCIE prototype operating mostly in homogeneous soil. Rule-based controllers often lack robustness and adaptability to varying conditions, and the influence of different soil types is not explored.
In~\cite{Autonomous2020Sotiropoulos}, an optimal control approach is proposed to minimize the distance between the rock and the bucket. A Gaussian Process (GP) modeling technique combined with an Unscented Kalman Filter is used to model the rock’s motion dynamics. The approach is implemented on a laboratory-scale UR10e manipulator equipped with a 3D-printed bucket to emulate an excavator equipped with a bucket, and its robustness to varying material properties is not examined.

\subsection{Contributions}\label{sec:Contributions}
This work introduces a RL–based control strategy for automating the rock capturing task with an excavator. Conventional model-based approaches are hindered by the complexity of unstructured environments and the highly dynamic interactions between the bucket, rock, and surrounding granular material. In contrast, our method leverages model-free learning and extensive randomization to achieve robust, generalizable, and efficient control. The main contributions are as follows:

\noindent\textbf{Data-driven control without explicit material modeling:} 
The proposed approach eliminates the need for analytical models of rock or soil properties. A PPO agent is trained entirely in simulation using excavator state variables and rock position information, directly generating joint speed commands for the excavator.

\noindent\textbf{Robustness through extensive domain randomization:}
Key parameters—including rock geometry, density, and mass, as well as the initial configurations of the rock, bucket, and goal—are randomized during training. This allows the controller to adapt effectively to unseen rock shapes and material properties during evaluation.

\noindent\textbf{Stable and efficient performance:}
The learned controller achieves a success rate of 0.8, comparable to human participants, while maintaining machine stability and avoiding excessive tilting. Once trained, the policy executes through a lightweight neural network forward pass, enabling real-time deployment in excavation tasks.


\section{Methodology}\label{sec:Methodolgy}
\begin{figure}[htbp] 
\centering
\includegraphics[width=0.75\linewidth]{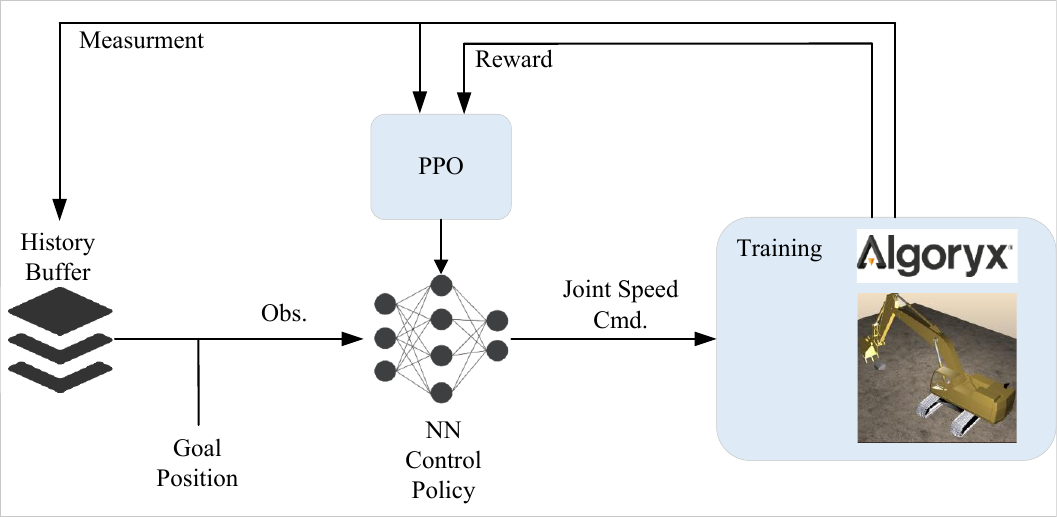}
\caption{The pipeline for training of the control policy.}
\label{fig:schematicMethod}
\end{figure}

In this section, an approach is described to learn a control policy entirety in simulation for automatic rock capturing using an excavator. In our framework, the term agent refers to the control policy, while the environment represents the excavator, rock, and terrain. A simple schematic of the proposed method is shown in Fig.~\ref{fig:schematicMethod}. First, the basic of RL is explained, then the modeling of the problem as a Markov Decision Process is elaborated.

\subsection{Goal-Conditioned Reinforcement Learning}
The problem is framed within the context of goal-conditioned RL, specifically as a finite horizon goal-conditioned Partially Observable Markov Decision Process (POMDP), represented by the tuple $(\mathcal{S},\Omega,\mathcal{G},\mathcal{A},O,P,r,H,\rho_0,\rho_g)$. At each time step $t$, the environment is in a state $s_t \in \mathcal{S}$, an observation $o_t \in \Omega$ is obtained, a goal $g_t \in \mathcal{G}$ is considered, and an action $a_t \in \mathcal{A}$ is executed. It is important to note that the goal remains fixed throughout the duration of an episode. Additionally, the observation model $O: \mathcal{S} \times \mathcal{A} \rightarrow \mathrm{Pr}(\Omega)$ defines the probability distribution over observations, the transition function $P: \mathcal{S} \times \mathcal{A} \rightarrow \mathrm{Pr}(\mathcal{S})$ characterizes the system's dynamics, and the reward function $r: \mathcal{S} \times \mathcal{A} \times \mathcal{G} \rightarrow \mathbb{R}$ specifies the feedback signal based on the state, action, and goal. Each episode is limited to a maximum length of $H$ time steps. The initial state and goal for each episode are sampled from the distributions $\rho_0$ and $\rho_g$, respectively~\cite{Takahiro2022Learning,Cong2022Reinforcement,Del2023Nonprehensile}. 

The objective is to learn a policy $\pi_{\theta}: \Omega \times \mathcal{G} \rightarrow \mathrm{Pr}(\mathcal{A})$ that maximizes the expected sum of discounted rewards $\mathbb{E}_\pi \left[ \sum_{t=0}^{H-1} \gamma^t r_t \right]$ where $\gamma\in (0,1)$ is a discount factor which trades off between current and future rewards. Our approach employs PPO to train a stochastic policy. PPO is a widely used on-policy RL algorithm, known for its effectiveness across diverse control tasks, such as locomotion~\cite{Emergence2017Nicolas} and in-hand manipulation~\cite{OpenAI2020Learning}. PPO aims to update the policy by maximizing a surrogate objective function that balances between improving the policy and maintaining stability. The key idea is to ensure that the new policy does not deviate too much from the old policy, preventing large, destabilizing updates~\cite{john2017Proximal}.

\subsection{Simulation Setup}\label{sec:SimulationSetup}
The simulation environment is built using AGX Dynamics\textsuperscript{\textregistered}, a high-fidelity physics engine designed for simulating complex mechanical systems. AGX Dynamics\textsuperscript{\textregistered} provides the accuracy and stability for simulating the intricate interactions between the excavator and granular terrain, enabling a reliable platform for developing autonomous excavation strategies~\cite{Servin2021MultiscaleTerrain,Aoshima2024Sim2Real}. The machine, based on the excavator CAT\textsuperscript{\textregistered}365, is a large excavator weighing 65,960$~kg$, with a maximum digging depth of $9.64~m$ and a bucket capacity of up to $3.8~m^3$. The excavator model in our simulation includes three linear actuators, which are actuated as motor-driven joints. 



The terrain is represented using a deformable 3D grid-based model, where properties such as mass, compaction, and soil type are discretely stored. Soil is deformed and moved via shovel objects, which interact with the terrain to simulate realistic digging and earthmoving behavior. These shovels convert static soil into dynamic mass aggregates, generating feedback forces based on soil mechanics theory, and allow for excavation, compaction, and soil redistribution~\cite{Servin2021Multiscale}. 

Moreover, a rock is introduced by creating its mesh, scaling the vertices to the desired size, and defining its collision geometry. The rock is assigned a material and then is added to the simulation along with the appropriate contact to ensure realistic interactions with the bucket, the terrain, and its particles. 

\subsection{Task Description}\label{sec:TaskDesciption}
\begin{figure}[htbp] 
    \centering
    \begin{subfigure}[b]{0.75\linewidth} 
        \centering
        \includegraphics[width=\linewidth]{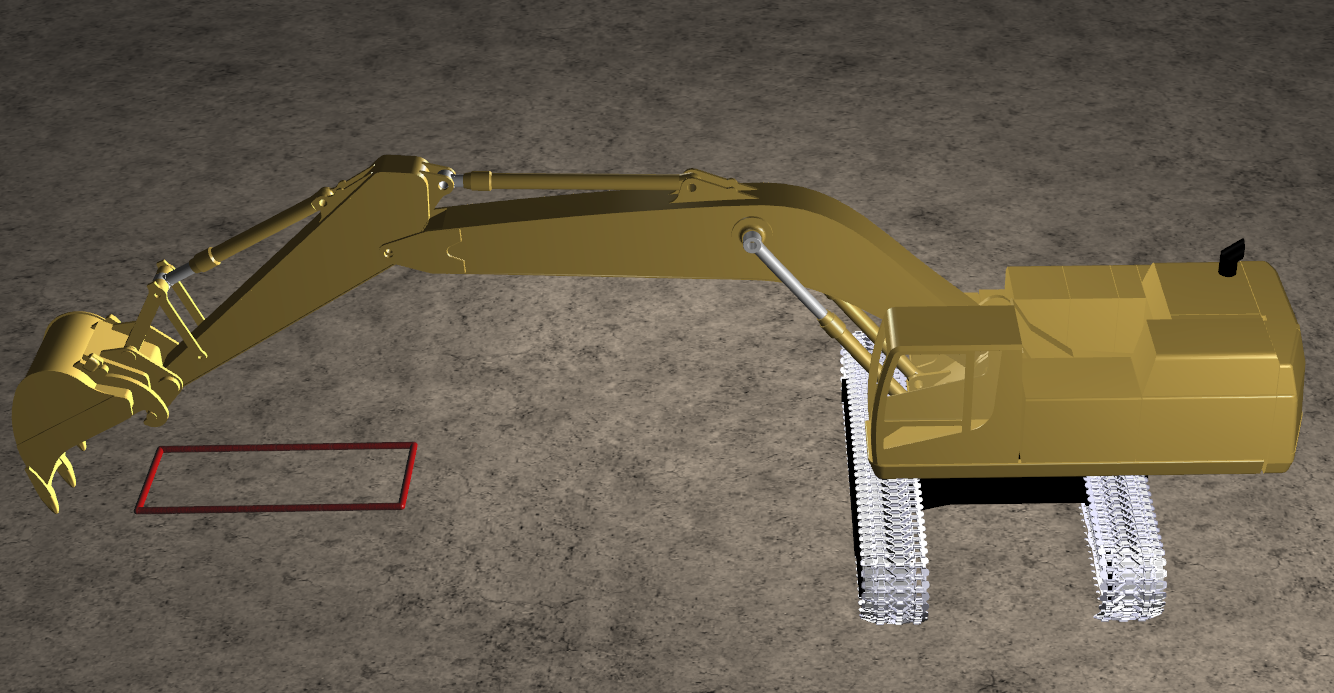}
        \caption{Maximum reach at ground level. }
        \label{fig:workspace_max}
    \end{subfigure}
    \hfill
    \begin{subfigure}[b]{0.75\linewidth} 
        \centering
        \includegraphics[width=\linewidth]{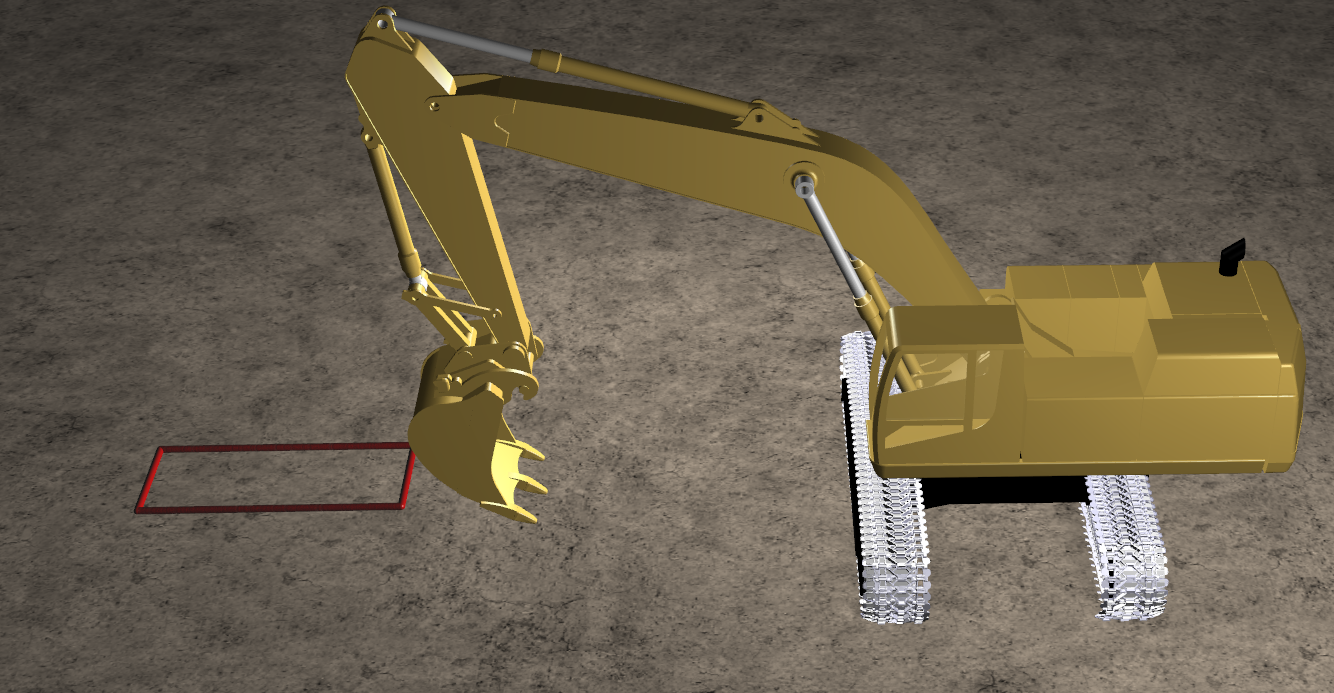} 
        \caption{Minimum reach at ground level.}
        \label{fig:workspace_min}
    \end{subfigure}
    \caption{Illustration of the maximum and minimum reach capabilities of the manipulator at ground level. The red boundary indicates the excavator's effective workspace where the rock capturing task should be performed.}
    \label{fig:workspace}
\end{figure}

The primary objective of the RL agent is to capture the rock using the bucket and move it to a designated goal location as quickly as possible. The task begins with the rock initialized at a location within the manipulator's effective workspace. Although the excavator's bucket can physically reach areas beyond the effective workspace, both closer to and farther from the machine, these regions are not considered part of the effective workspace because the bucket cannot perform the rock capturing task effectively there. In real-world scenarios, placing the rock too close or too far from the excavator is unrealistic. For successful operation, the rock must be positioned within the machine’s effective workspace. The effective workspace of the boom, arm, and bucket is illustrated by the red boundary in Fig.~\ref{fig:workspace}.

The excavator is controlled by generating speed commands to the bucket, arm, and boom joints. Although the excavator model includes components such as the cabin hinge joint and tracks, these are not actuated in this task, as three joints of manipulator are sufficient for the rock capturing.
In this task, maintaining stability is critical, as improper interaction between the bucket and the terrain can cause the bucket to get stuck, potentially leading to tilting of the excavator.
Another critical aspect of the task is the interaction between the bucket and the rock. If the bucket approaches the rock from an incorrect Point of Attack (PoA) or with excessive speed, the rock may be thrown, which is undesirable and potentially dangerous. At the end of the task, the rock should be securely inside the bucket and positioned close enough to the goal position.

\subsection{Episode Initialization}\label{sec:Initialization}
At the beginning of each training episode, the environment is initialized in a randomized state. The randomized parameters are listed in Table~\ref{tab:Randomization}.
\begin{table}[ht]
    \centering
    \caption{Randomized parameters at the beginning of each training episode.}
    \label{tab:Randomization}
    \begin{tabular}{lc}
        \toprule
        Parameter & Distribution  \\
        \midrule
        Goal position & $ \mathcal{N} \left(\boldsymbol{\mu}=
        \begin{bmatrix}
           -7.0 \\
           1.5 
         \end{bmatrix},\boldsymbol{\Sigma}=
        \begin{bmatrix}
        0.1^2  & 0 \\
        0 & 0.1^2
        \end{bmatrix}\right)$\\
        Rock density &$\mathcal{N} \left(\mu=2000.0, \sigma^2 = 85.0^2 \right)$ \\
        Rock geometry & $\mathcal{U}\{I, II\}$ \\
        Rock position ($x$-axis) &
        $\mathcal{U}(-11.5,-8.0)$ \\
        \bottomrule
    \end{tabular}
\end{table}

\begin{figure}[htbp] 
\centering
\includegraphics[width=0.75\linewidth]{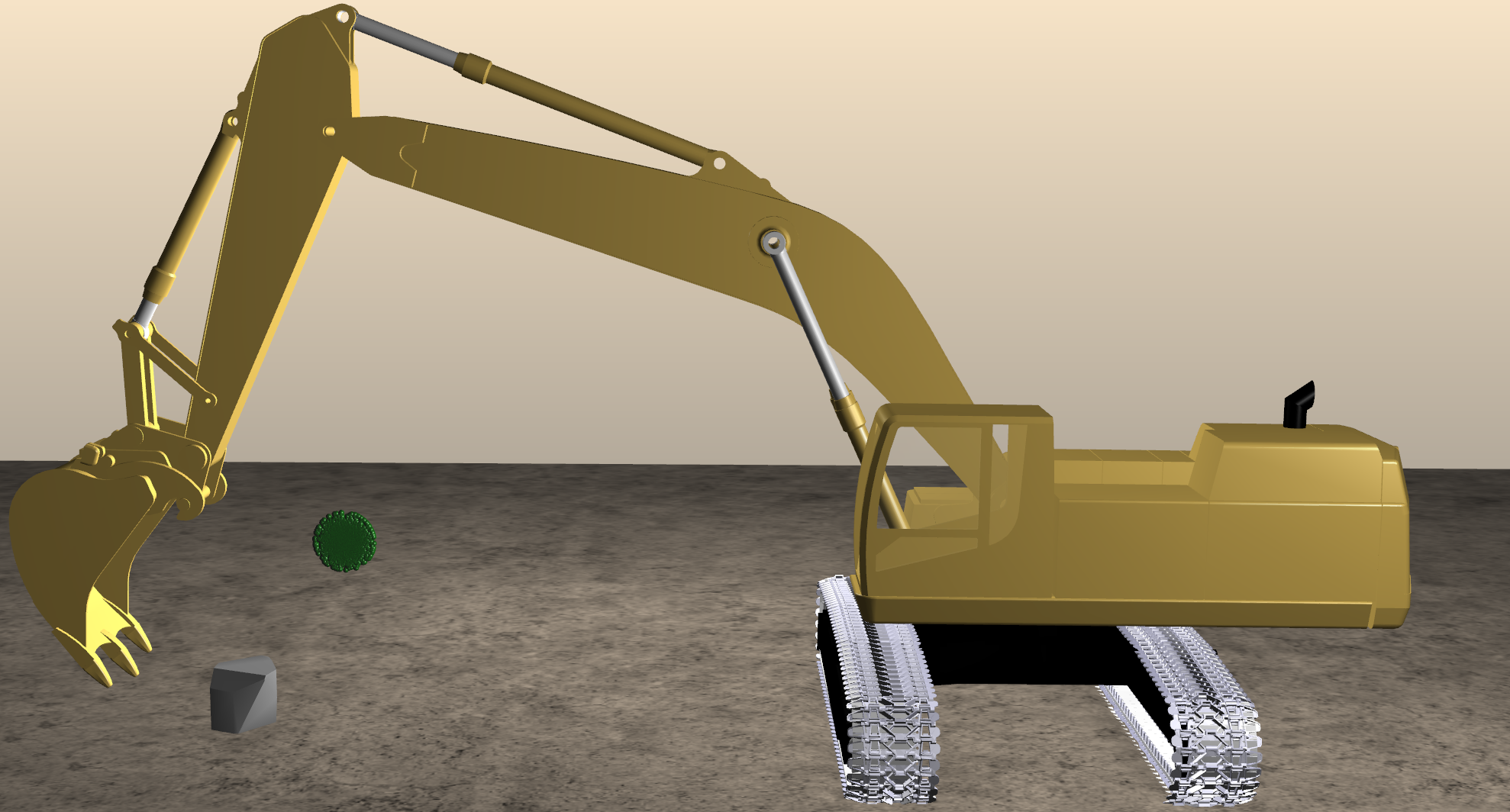}
\caption{Illustration of randomly sampled goal positions during training. The samples are drawn from the bivariate normal distribution $\mathcal{N} \left(\boldsymbol{\mu},\boldsymbol{\Sigma}\right)$ and constrained within a circular region of radius $0.3~m$.}
\label{fig:GoalPositionsdist}
\end{figure}

The goal position is sampled from a 2D normal distribution (also known as a bivariate normal distribution) $\mathcal{N} \left(\boldsymbol{\mu},\boldsymbol{\Sigma}\right)$, where the mean $\boldsymbol{\mu}$ and standard deviation $\boldsymbol{\Sigma}$ values are presented in Table~\ref{tab:Randomization}. To ensure feasibility, the sampled position is constrained within a circle of radius $0.3~m$. If a sample falls outside this boundary, it is projected onto the circle's edge. For better understanding, a visualization of randomly sampled goal positions is shown in Fig.~\ref{fig:GoalPositionsdist}.

\begin{figure}[htbp] 
    \centering
    \begin{subfigure}[b]{0.4\linewidth}
        \centering
        \includegraphics[width=\textwidth]{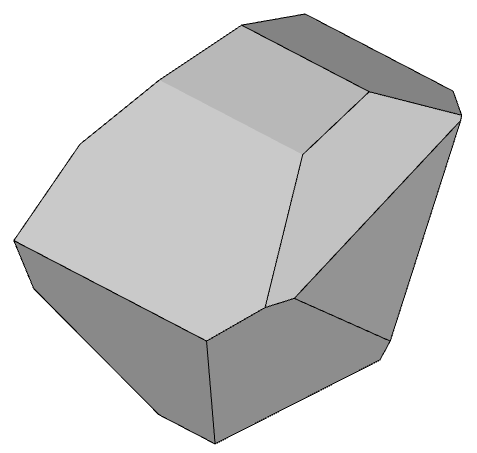}
        \caption{Geometry of rock~I.}
        \label{fig:rockGeos_2}
    \end{subfigure}
    \begin{subfigure}[b]{0.4\linewidth}
        \centering
        \includegraphics[width=\textwidth]{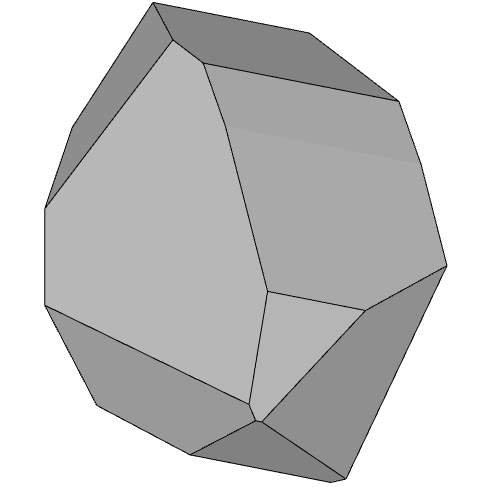} 
        \caption{Geometry of rock~II.}
        \label{fig:rockGeos_3}
    \end{subfigure}
    \caption{The geometries of the rocks used during training. Each episode randomly selects one of the two rock meshes.}
    \label{fig:rockGeos}
\end{figure}
A rock is added to the environment with a randomly selected geometry from two different mesh models, as illustrated in Fig.~\ref{fig:rockGeos}. The randomization of geometry and density results in significant randomization of the rock mass. Further implementation details and parameter specifications for the environment initialization are provided in~\ref{appendix:A}.

\subsection{Observation}\label{sec:Observation}
At each time step $t$, the policy receives an observation $o_t$ that includes the position, speed, and force of the three prismatic joints; the $x$- and $z$-coordinates of the goal position, the position of the rock's Center of Mass (CoM), and the center position of the bucket. All quantities are expressed in the base frame. The base frame is located beneath the cabin at the excavator’s swing pin. Its $x$-axis points toward the rear of the machine (i.e., opposite the direction the excavator faces), its $z$-axis points upward, and the $y$-axis is defined according to the right-hand rule. The motion of the excavator's manipulator is confined to the $x$–$z$ plane of the base frame.
The observation also contains the roll $\phi$ and pitch $\theta$ angles of the base frame. To maintain the stability and avoid tilting of the excavator during the operation, the roll $\phi$ and pitch $\theta$ angles of the base frame play a key role.
However, these observations do not capture other aspects of the environment state, such as the frictional contact forces, the contact points, or the mass and geometric properties of the rock. These unobserved variables can affect the dynamics of interaction and task success. The observations are summarized in Table~\ref{tab:actobs}.
\begin{table}[htbp]
    \centering
    \begin{tabular}{lcc}
    \toprule
    Observations $o_t$ & Notation  & Unit \\
    \midrule
    Joint positions & $(q_{\text{boom}}, q_{\text{arm}}, q_{\text{bucket}})$ & $m$  \\
    Joint speeds & $(v_{\text{boom}}, v_{\text{arm}}, v_{\text{bucket}})$ & $m/s$ \\
    Joint forces & $(f_{\text{boom}}, f_{\text{arm}}, f_{\text{bucket}})$ & $k N$  \\
    Bucket position & $(x_{\text{bucket}},z_{\text{bucket}})$ & $m$ \\
    Rock position & $(x_{\text{rock}},z_{\text{rock}})$ & $m$  \\
    Goal position & $(x_{\text{goal}},z_{\text{goal}})$ & $m$  \\
    Cabin pitch and roll angles & $(\theta,\phi)$ & $rad$ \\
    \bottomrule
    \end{tabular}
    \caption{Observations used by the policy at each time step. The observation vector $o_t$ has a total dimension of $17$. Units and notation for each observation are shown in the table.}
    \label{tab:actobs}
\end{table}
In addition, to accelerate convergence, the observations are normalized to the range $[-1,1]$ using their minimum and maximum values.



\subsection{Goal and Action}\label{sec:GoalAction}
The policy is trained to control the excavator to capture a rock and move it toward a designated goal position in the base frame. The rock is considered close enough to the goal if the horizontal and vertical distances between the rock's CoM and the goal position are each less than proximity threshold $\delta_{\text{prox}}$, meaning the proximity condition defined in Eq.~\eqref{eq:TargetCloseEnough} is satisfied:
\begin{equation}\label{eq:TargetCloseEnough}
\mathbb{C}_{\text{proximity}} = 
\begin{cases}
1, & \begin{aligned}
      if~|x_{\text{rock}} - x_{\text{goal}}| < \delta_{\text{prox}}~and \\
      |z_{\text{rock}} - z_{\text{goal}}| < \delta_{\text{prox}}
    \end{aligned} \\
0, & otherwise
\end{cases}
\end{equation}
In addition, another condition regarding the tilting of the excavator is required to obtain the goal condition. The roll $\phi$ and pitch $\theta$ angles of the cabin should be less than tilt threshold $\delta_{\text{tilt}}$, meaning the tilting condition defined in Eq.~\eqref{eq:tiltingCondition} is satisfied:
\begin{equation}\label{eq:tiltingCondition}
\mathbb{C}_{\text{tilting}} = 
\begin{cases}
1, & if~|\phi| <  \delta_{\text{tilt}}~and~|\theta| <  \delta_{\text{tilt}} \\
0, & otherwise
\end{cases}
\end{equation}
The combination of proximity condition $\mathbb{C}_{\text{proximity}}$ and tilting condition $\mathbb{C}_{\text{tilting}}$ is expressed as the goal condition $\mathbb{C}_{\text{goal}}$:
\begin{equation}\label{eq:goalCondition}
\mathbb{C}_{\text{goal}} = 
\begin{cases}
1, & if~\mathbb{C}_{\text{proximity}}~and~\mathbb{C}_{\text{tilting}} \\
0, & otherwise
\end{cases}
\end{equation}
Given a goal $g_t$ and an observation $o_t$, the policy takes an action $a_t = \left[v_{\text{boom}},v_{\text{arm}}, v_{\text{bucket}}\right]^T$, which consists of the joint speed of the boom, arm, and bucket. The trained policy generates normalized speed commands in the range of $[-1,1]$, which are then scaled by the maximum speed limits of each joint. The maximum speed commands for the boom, arm, and bucket joints are $0.3~m/s$, $0.3~m/s$, and $0.2~m/s$, respectively.

\subsection{Reward}\label{sec:Reward}
Designing an effective reward function is crucial for enabling fast and stable learning, especially in environments with sparse rewards, where the agent must explore extensively before discovering rewarding behaviors. To address this challenge, a guiding reward formulation is adopted, inspired by prior works~\cite{vasan2024revisitingsparserewardsgoalreaching,OpenAIGym}. Unlike sparse rewards, which provide feedback only upon satisfying the goal conditions, the guiding reward offers intermediate feedback by encouraging the agent to move toward the target and to maintain the desired state once reached.
In our case, the agent receives a reward $r_t$ at each time step, composed of two components: a guidance reward $r_{\text{guidance}}$, which incentivizes progress toward the goal state, and a goal reward $r_{\text{goal}}$, which encourages the agent to remain within the goal region once it is reached:
\begin{equation}\label{eq:reward}
r_t = r_{\text{guidance}} + r_{\text{goal}}
\end{equation}
This reward shaping approach facilitates both exploration and convergence. However, it is important to note that reward functions designed through trial-and-error can lead to reward overfitting, where the learned behavior becomes overly tailored to a specific algorithm or training scenario~\cite{vasan2024revisitingsparserewardsgoalreaching}. To mitigate this risk, the guiding reward are deliberately kept abstract and general, motivating the agent to reach and maintain the goal state, without relying on detailed task-specific heuristics. A fixed episode length is long enough to ensure sufficient time for goal achievement while promoting efficient learning.
The guidance reward $r_{\text{guidance}}$ is formulated as follows:
\begin{align}\label{eq:guidanceReward}
r_{\text{guidance}} =\ & 
- \frac{1}{w_1} \left( x_{\text{rock}} - x_{\text{goal}} \right)^2 \notag \\
& - \frac{1}{w_2} \left( z_{\text{rock}} - z_{\text{goal}} \right)^2 \notag \\
& - \frac{1}{w_3} \left\| a_t \odot f_t \right\|_2^2 \notag \\
& - \frac{1}{w_4} \left\| a_t - a_{t-1} \right\|_2^2 \notag \\
& - \frac{1}{w_5} \left( \theta^2 + \phi^2 \right)
\end{align}
where $f_t = [f_{\text{boom}}, f_{\text{arm}}, f_{\text{bucket}}]^T$ denotes the joint forces. The first and second terms penalize the distance between the rock’s CoM and the goal in the $x$- and $z$-axes. The third term penalizes energy usage by minimizing the product of joint forces and speeds, the fourth term penalizes unnecessary or excessive variations in control input (joint speeds) to promote smooth movements, and the last term penalizes cabin tilt to ensure the excavator remains stable. The notation $\left\| \cdot \right\|_2$ denotes the Euclidean norm ($2$-norm), and  $\odot$ indicates the element-wise product. The weights $w_i,~i \in \{1,2,3,4,5\}$ serve to normalize the associated penalty terms and control the trade-offs among the different components. 
The goal reward \( r_{\text{goal}} \) is defined as:
\begin{equation}\label{eq:goalreward}
r_{\text{goal}} = 
\begin{cases}
5, & if~\mathbb{C}_{\text{goal}} \\
0, & otherwise
\end{cases}
\end{equation}
Moreover, an episode truncates if the rock becomes inaccessible to the bucket. The truncate condition is defined as:
\begin{equation}\label{eq:terminalCondition}
\mathbb{C}_{\text{truncate}} = 
\begin{cases}
1, & if~|y_{\text{rock}}| > y_{truncate}~or~x_{\text{rock}} < x_{truncate} \\
0, & otherwise
\end{cases}
\end{equation}
where $y_{\text{rock}}$ is the position of the rock’s CoM in $y$-axis in the base frame. The $\mathbb{C}_{\text{truncate}}$ indicates whether the rock lies outside the excavator’s effective workspace: $x_{\text{rock}}<x_{truncate}$ means it is too far, and $|y_{\text{rock}}|>y_{truncate}$ means it is outside the $x-z$ plane of manipulator motion.
The parameters used in the reward function and conditions are listed in Table~\ref{tab:task_parameters}. Finally, an episode terminates if the episode length reaches to the maximum episode length $H$.
\begin{table}[htbp]
\centering
\caption{Parameters used in the guidance reward $r_{\text{guidance}}$ and task conditions.}
\label{tab:task_parameters}
\begin{tabular}{l c c}
\toprule
Parameter & Value & Unit \\
\midrule
$w_1$ & $13.0$ & -- \\
$w_2$ & $8.0$ & -- \\
$w_3$ & $3.0 \times 200.0^2$ & -- \\
$w_4$ & $12.0$ & -- \\
$w_5$ & $1.0$ & -- \\
$\delta_{\text{prox}}$ & $0.2$ & $m$ \\
$\delta_{\text{tilt}}$ & $0.1$ & $rad$ \\
$x_{truncate}$ & $-13.0$ & $m$ \\
$y_{truncate}$ & $1.0$ & $m$ \\
\bottomrule
\end{tabular}
\end{table}



\subsection{Training}\label{sec:Trainig}
The agent is trained using the PPO algorithm~\cite{john2017Proximal} implemented in the Stable-Baselines3 library~\cite{Stable2021Antonin}. Separate neural networks are used to represent the policy and value functions, each receiving identical inputs and having a linear output layer.
The training hyperparameters are provided in Table~\ref{tab:ppo_hyperparameters}.
\begin{table}[htbp] 
\centering
\caption{PPO training hyperparameters.}
\label{tab:ppo_hyperparameters}
\begin{tabular}{ll}
\toprule
Parameter                 & Value                                         \\
\midrule
Policy hidden layers      & $128 \times 128$, tanh \\
Value hidden layers       & $128 \times 128$, tanh                      \\
Discount factor $\gamma$  & $0.99$                                          \\
Max. episode length       & $500 (@60Hz) \approxeq 8.3~s$                                       \\
Entropy coefficient       & $3\times 10^{-4}$                                          \\
Learning rate             & $3\times 10^{-4}$                                       \\
Value function coefficient & $0.5$                                           \\
Max. grad norm            & $0.5$                                           \\
GAE $\lambda$              & $0.95$                                          \\
Mini-batches              & $128$                                             \\
Optimization epochs       & $4$                                             \\
Clip range                & $0.2$                                           \\
\bottomrule
\end{tabular}
\end{table}
All training and experience collection are performed on a Linux workstation running Ubuntu $22.04$ LTS. The system is equipped with an Intel Xeon E$5$-$1650$ v$2$ CPU ($6$ physical cores, $12$ threads @ $3.5$~GHz), an NVIDIA GeForce RTX $4070$ GPU with $12$~GB of VRAM, and $64$~GB of system RAM. The policy is trained for $15\times10^{6}$ time steps, which lasts around $17~h$.



\section{Results}\label{sec:results}
In this section, the results of the proposed method are presented. First, the cumulative reward obtained during policy training is illustrated. The training setup is extensively described in Sections~\ref{sec:SimulationSetup} and \ref{sec:TaskDesciption}, where the rock’s geometry (shown in Fig.~\ref{fig:rockGeos}) and density are randomized, its initial position is sampled within the excavator’s effective workspace, and the goal position is drawn from a normal distribution. The soil is modeled as cohesive dirt, whose physical parameters are listed in Table~\ref{tab:dirtMaterial}. Then, the trained policy is evaluated across four distinct scenarios: (I) under conditions similar to the training environment, (II) with rocks different from those used during training, (III) with soil properties that differ from the training conditions, and (IV) in comparison with human participants using a game-like control interface.

\begin{figure}[htbp] 
\centering
\includegraphics[width=0.7\linewidth]{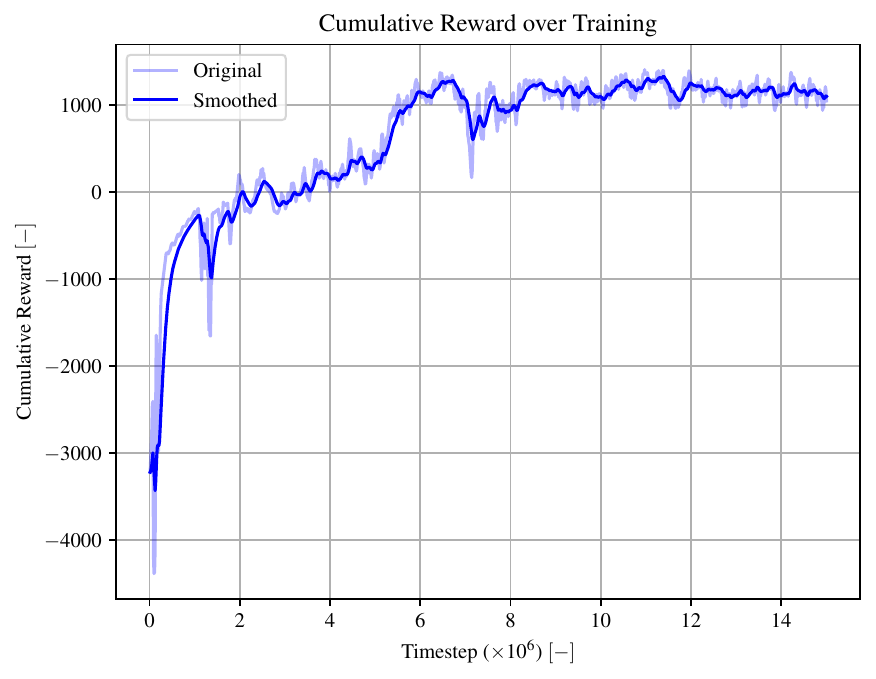}
\caption{Cumulative reward obtained during training. To improve interpretability, a smoothing function (Eq.~\eqref{eq:smoothFun}) with $w_s = 0.9$ is applied to the cumulative reward. The original values are shown with reduced opacity for visual reference.}
\label{fig:cumlativeReward}
\end{figure}

The cumulative reward curve, shown in Fig.~\ref{fig:cumlativeReward}, represents the mean episodic cumulative reward, averaged over the most recent $100$ episodes.
\begin{figure}[htbp] 
\centering
\includegraphics[width=0.7\linewidth]{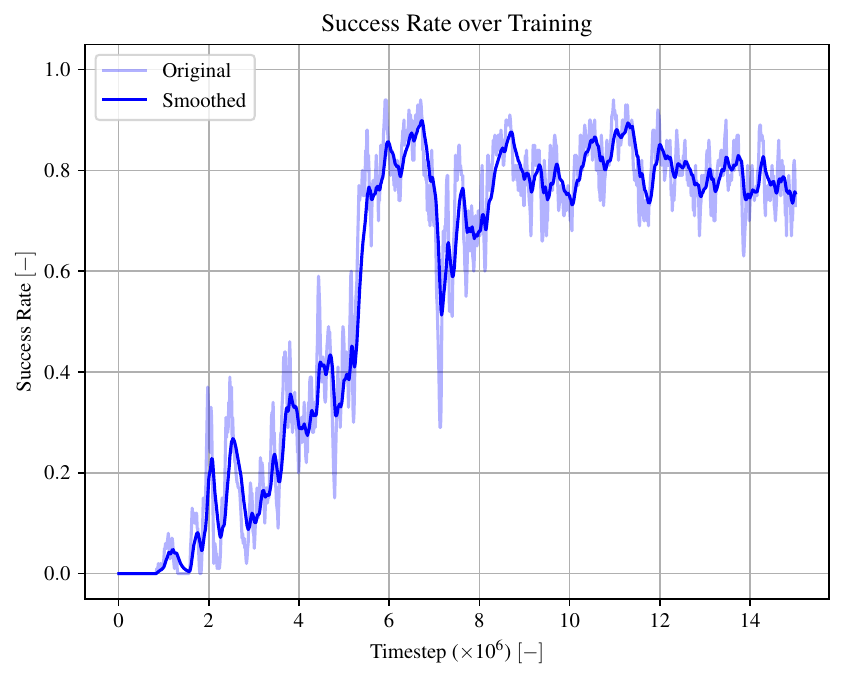}
\caption{Success rate obtained during training. To improve interpretability, a smoothing function (Eq.~\eqref{eq:smoothFun}) with $w_s = 0.9$ is applied to the success rate. The original values are shown with reduced opacity for visual reference.}
\label{fig:SuccessRate}
\end{figure}
To improve interpretability, the cumulative reward is smoothed using Eq.~\eqref{eq:smoothFun}:
\begin{equation}\label{eq:smoothFun}
s(t) = \left\{
\begin{array}{ll}
x(0), & if~t = 0 \\
w_s s(t-1) + (1-w_s) x(t) , & if~t>0
\end{array}
\right.
\end{equation}
where $x(t)$ is the original data at time step $t$, $s(t)$ is the smoothed value at time step $t$, and $w_s \in [0,1]$ is the smoothing weight. Larger values of $w_s$ result in greater smoothing, making the output less sensitive to recent changes in the data. Lower values of $w_s$ provide less smoothing and give greater weight to recent observations.
As shown in Fig.~\ref{fig:cumlativeReward}, the cumulative reward increases until approximately $8$ million time steps, after which it plateaus, indicating that the policy has converged.
The success rate is presented in Fig.~\ref{fig:SuccessRate}. This value represents the fraction of episodes during a rollout (a window of recent $100$ episodes) in which the agent successfully reaches the goal. An episode is considered successful if the goal condition $\mathbb{C}_{\text{goal}}$ defined in Eq.~\eqref{eq:goalCondition} is satisfied. At the end of the training, the success rate is around $0.8$ which shows satisfactory performance for this complex task.

\subsection{Evaluation Under Training Conditions}\label{sec:EvaTrainingCondition}
In this section, the learned policy is evaluated over $10$ episodes under conditions similar to those used during training. The success rate, along with the mean and standard deviation of the cumulative reward of the agent across different evaluation scenarios, is summarized in Table~\ref{tab:SRAndMeanStdCR}.
\begin{table}[htbp]
    \centering
    \caption{Success rates and mean and standard deviation of cumulative reward over $10$ episodes for each evaluation scenario.}
    \label{tab:SRAndMeanStdCR}
    \begin{tabular}{lcc}
        \toprule
        Evaluation Scenario & Success Rate & Cumulative Reward\\
        \midrule
        Training condition & $0.9$ & $1428.47\pm611.13$ \\
        Unseen rocks & $0.8$ & $1195.84\pm745.17$ \\
        Unseen materials & $0.7$ & $952.04\pm779.10$ \\
        \bottomrule
    \end{tabular}
\end{table}

\begin{figure}[htbp] 
\centering
\includegraphics[width=0.7\linewidth]{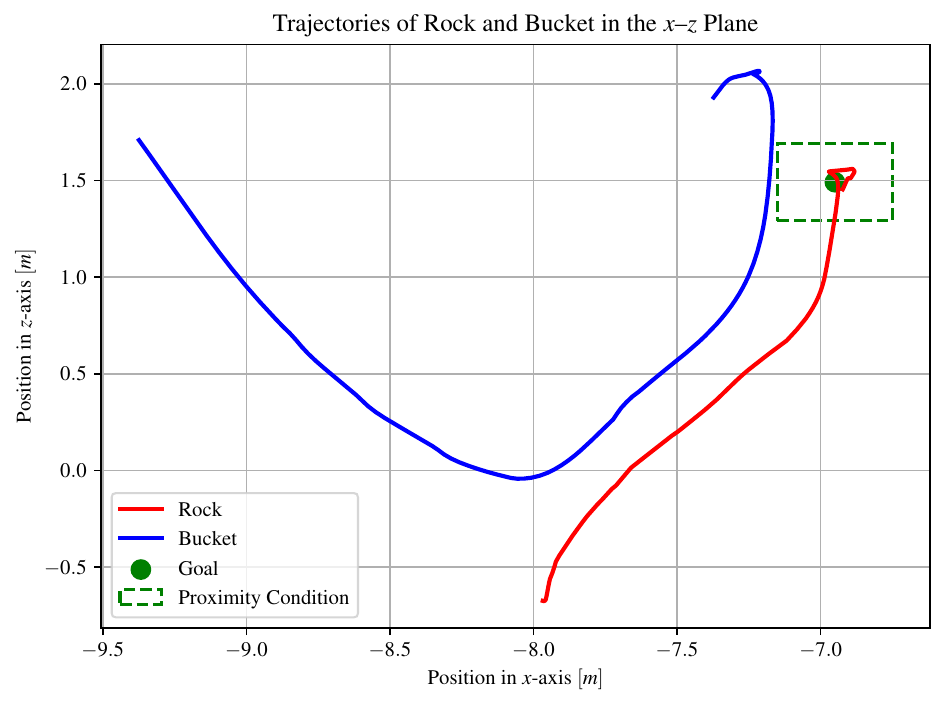}
\caption{Trajectories of the rock and the bucket in the $x$-$z$ plane during a successful episode under training conditions. The green square denotes the proximity condition.}
\label{fig:TrajectoryTrainingCondition}
\end{figure}

To facilitate deeper analysis, one successful episode is randomly selected for detailed examination. The trajectories of the rock and the bucket are illustrated in Fig.~\ref{fig:TrajectoryTrainingCondition}, showing that the rock successfully reaches the goal and remains within its proximity.
\begin{figure}[htbp] 
\centering
\includegraphics[width=0.7\linewidth]{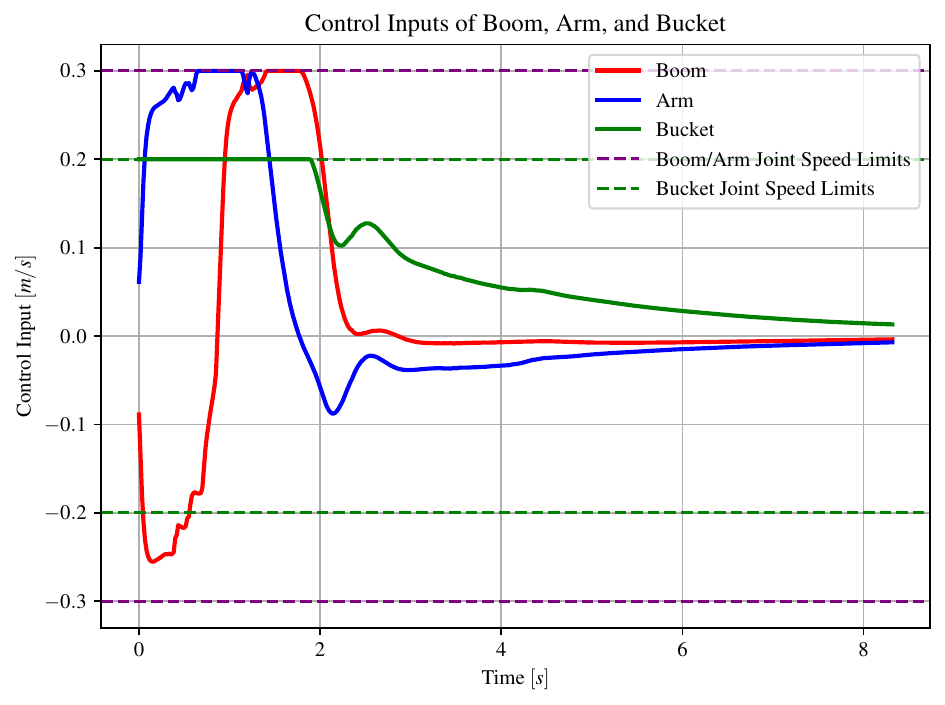}
\caption{Joint speed commands for the bucket, arm, and boom throughout a successful episode under training conditions.}
\label{fig:ControlInputTrainingCondition}
\end{figure}
\begin{figure}[htbp] 
\centering
\includegraphics[width=0.7\linewidth]{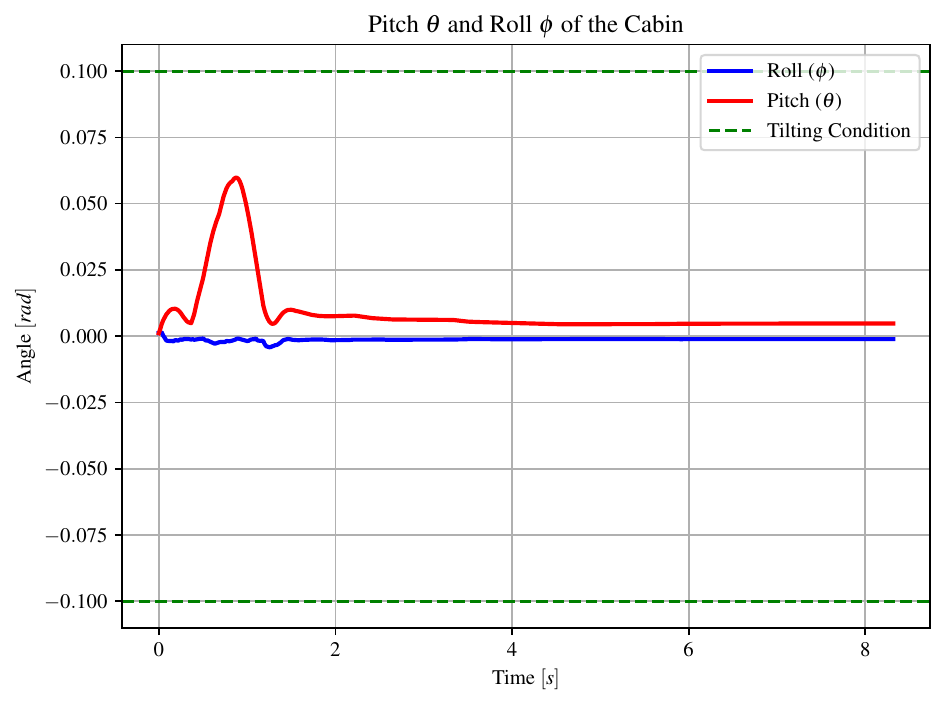}
\caption{Pitch $\theta$ and roll $\phi$ angles of the cabin during a successful episode under training conditions. The valid range defined by the tilting condition $\mathbb{C}_{\text{tilting}}$ is indicated by green dashed lines.}
\label{fig:tiltingTrainingCondition}
\end{figure}
The control inputs, defined as the joint speeds of the boom, arm, and bucket, are shown in Fig.~\ref{fig:ControlInputTrainingCondition}. In the simulation, the excavator immediately executes the commanded joint speeds, so the control inputs directly match the actual joint speeds. It is worth noting that, unlike the simulation, a real excavator exhibits delays between commanded and measured joint speeds, which introduces a sim-to-real gap that must be considered for deployment. Toward the end of the episode, the joint speeds gradually converge to zero, which helps stabilize the bucket and the rock near the goal location. A smooth deceleration is crucial, since abruptly driving the joints to zero could cause the rock to be thrown out of the bucket due to inertial effects, potentially falling to the ground or even being projected toward the cabin, which would be unsafe in real-world operation. As observed, the signals exhibit no jerky or excessive actions, suggesting that the generated control inputs are feasible for real-world operations.
Finally, the cabin’s pitch $\theta$ and roll $\phi$ angles are shown in Fig.~\ref{fig:tiltingTrainingCondition}. Both angles remain within the valid range defined by the tilting condition $\mathbb{C}_{\text{tilting}}$.


\subsection{Evaluation Using Unseen Rock Geometries}\label{sec:UnseenRock}
\begin{figure}[htbp] 
    \centering
    \begin{subfigure}[b]{0.4\linewidth}
        \centering
        \includegraphics[width=\textwidth]{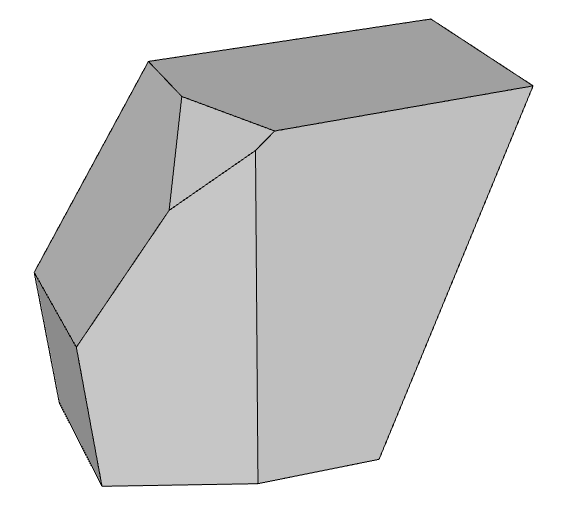}
        \caption{Geometry of rock~III.}
        \label{fig:rockGeos_4}
    \end{subfigure}
    \begin{subfigure}[b]{0.4\linewidth}
        \centering
        \includegraphics[width=\textwidth]{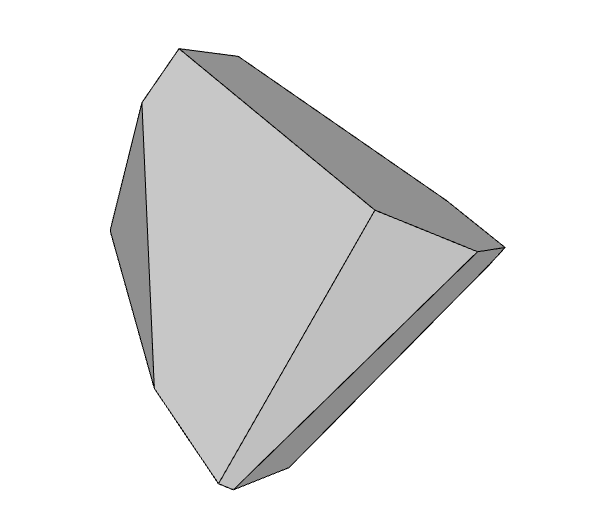} 
        \caption{Geometry of rock~IV.}
        \label{fig:rockGeos_5}
    \end{subfigure}
    \caption{The geometries of the rocks used in the unseen rock evaluation scenario. Each episode randomly selects one of the two rock meshes.}
    \label{fig:rockGeosTest}
\end{figure}

In this section, the learned policy is evaluated over 10 episodes using two unseen rock geometries to assess the policy’s generalization capabilities. The shapes of these rocks are shown in Fig.~\ref{fig:rockGeosTest}.
The success rate, along with the mean and standard deviation of the cumulative reward for this scenario, is summarized in Table~\ref{tab:SRAndMeanStdCR}. A significant reduction in both metrics compared to the training conditions is not observed, demonstrating that the policy does not overfit to specific rock geometries, densities, or masses.

\begin{figure}[htbp] 
\centering
\includegraphics[width=0.7\linewidth]{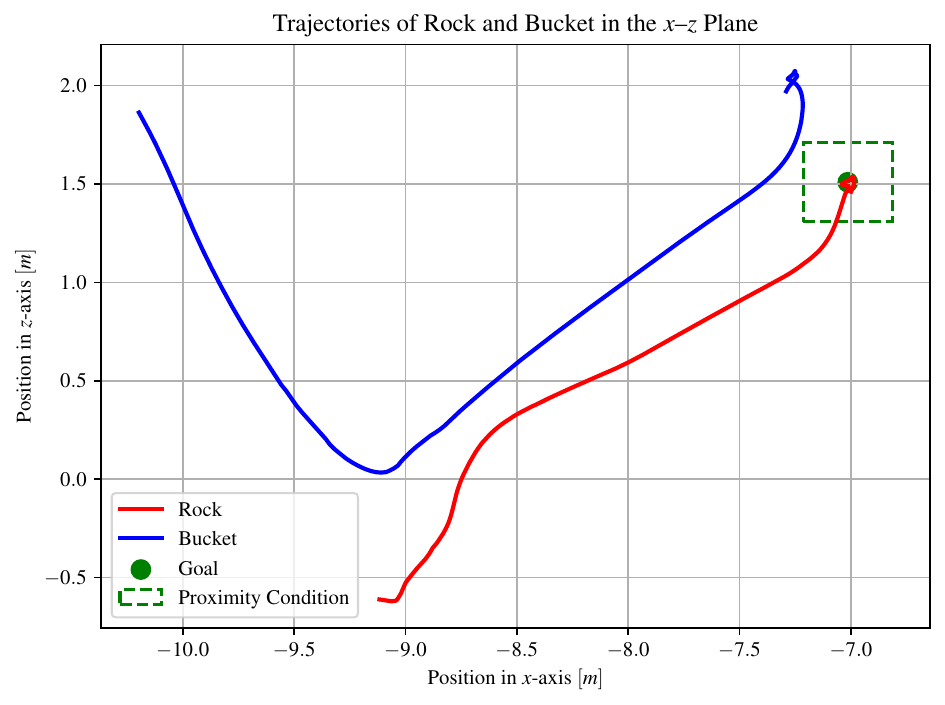}
\caption{Trajectories of the rock and bucket in the $x$-$z$ plane during a successful episode in unseen rock evaluation scenario. The green square denotes the proximity condition.}
\label{fig:TrajectoryRockTestCondition}
\end{figure}

\begin{figure}[htbp] 
\centering
\includegraphics[width=0.7\linewidth]{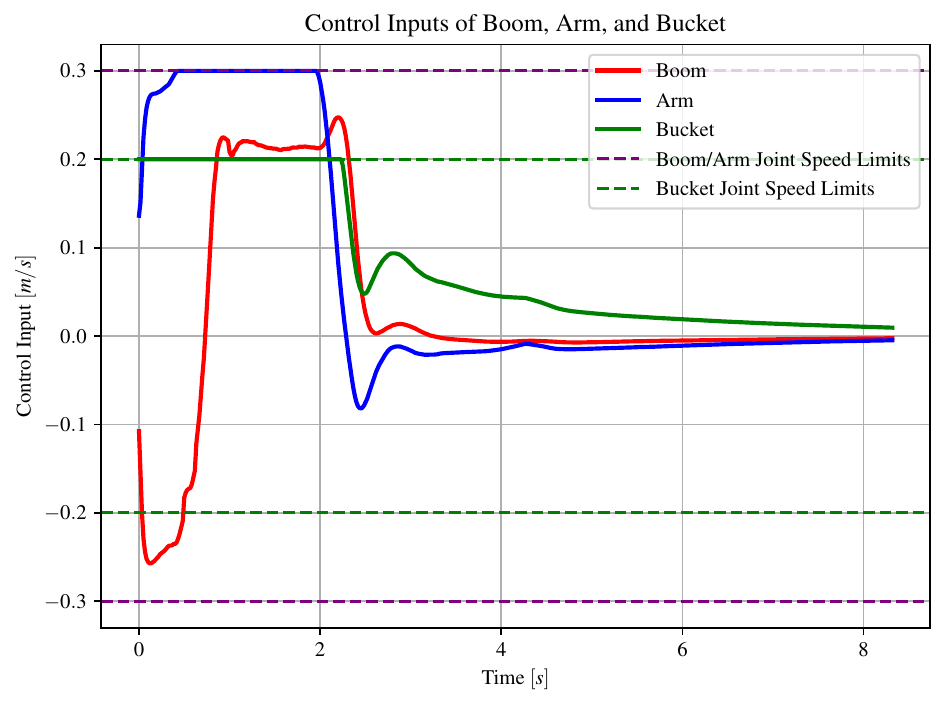}
\caption{Joint speed commands for the bucket, arm, and boom throughout a successful episode in unseen rock evaluation scenario.}
\label{fig:ControlInputRockTestCondition}
\end{figure}

\begin{figure}[htbp] 
\centering
\includegraphics[width=0.7\linewidth]{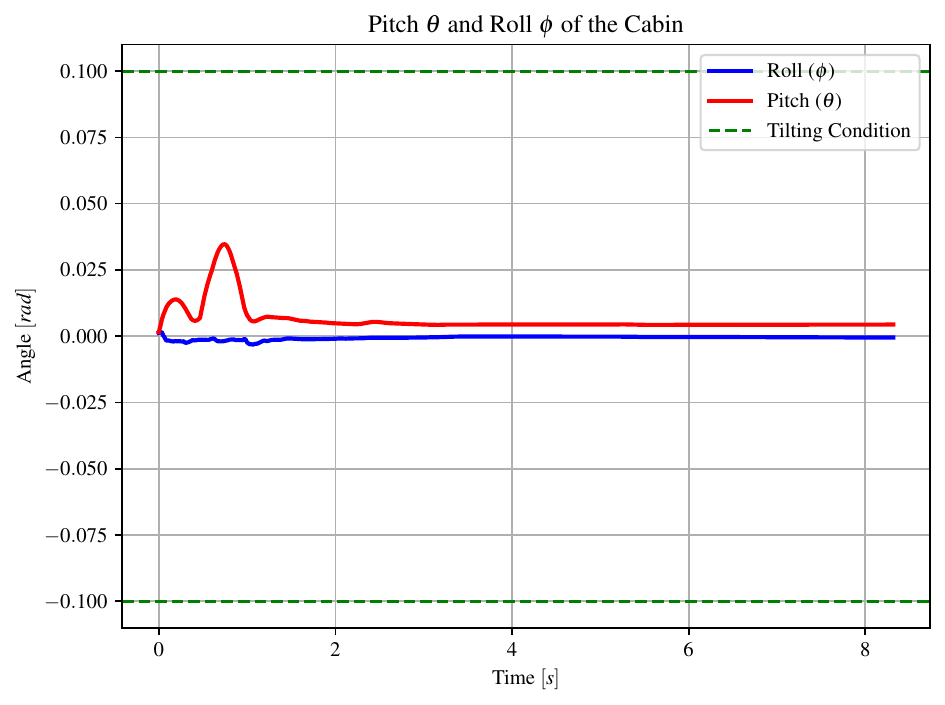}
\caption{Pitch $\theta$ and roll $\phi$ angles of the cabin during a successful episode in unseen rock evaluation scenario. The valid range defined by the tilting condition $\mathbb{C}_{\text{tilting}}$ is indicated by green dashed lines.}
\label{fig:tiltingRockTestCondition}
\end{figure}

To provide further insight, one successful episode is randomly selected for detailed analysis. 
The trajectories of the rock and bucket in the $x$-$z$ plane are illustrated in Fig.~\ref{fig:TrajectoryRockTestCondition}. The rock clearly reaches and remains within the goal region (highlighted by the green square). Notably, the movement appears near-optimal, with an almost direct trajectory and minimal unnecessary motion.
Figure~\ref{fig:ControlInputRockTestCondition} displays the control inputs (joint speed commands) for the bucket, arm, and boom. After approximately $2~s$, the inputs reduce to stabilize the rock near the goal position. The absence of abrupt or excessive commands supports the real-world feasibility of the learned policy.
Finally, the cabin’s pitch $\theta$ and roll $\phi$ angles are plotted in Fig.~\ref{fig:tiltingRockTestCondition}. Only minor variations are observed, all of which remain within the allowable tilting range defined by $\mathbb{C}_{\text{tilting}}$.

\subsection{Evaluation Using Unseen Material Properties}\label{sec:UnseenSoil}
In this section, the controller is evaluated over $10$ episodes using sand as material to assess the policy’s generalization to varying material properties. The key physical parameters defining the sand material are provided in~\ref{appendix:B}. The success rate, as well as the mean and standard deviation of the cumulative reward for this scenario, are reported in Table~\ref{tab:SRAndMeanStdCR}. The controller’s performance in this setting is comparable to that observed under the training conditions and with unseen rock geometries, indicating that the learned policy is not sensitive to changes in material properties.

\begin{figure}[htbp] 
\centering
\includegraphics[width=0.7\linewidth]{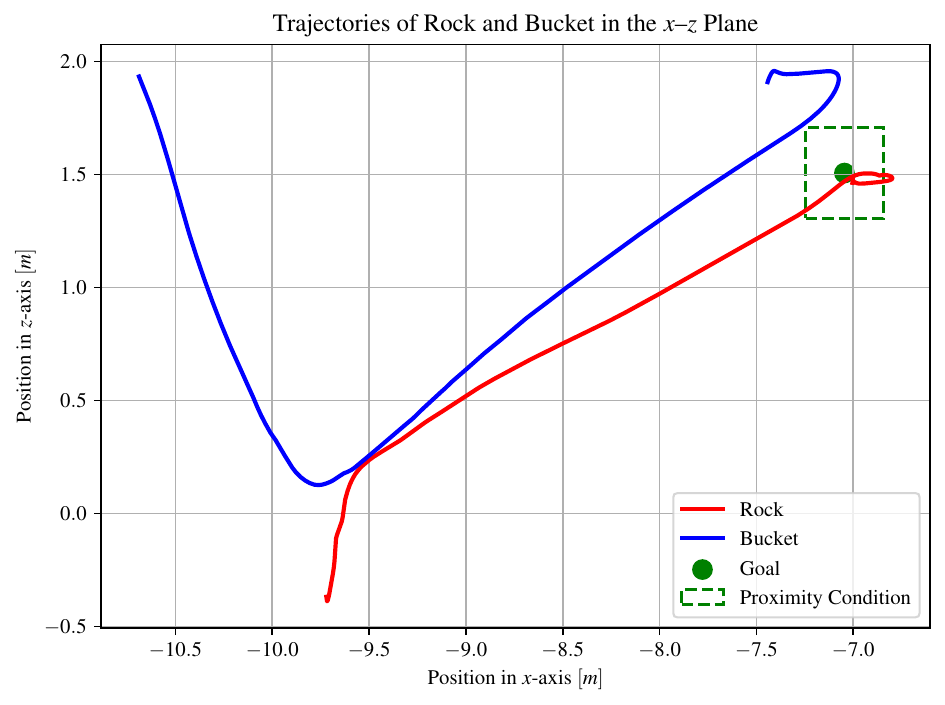}
\caption{Trajectories of the rock and bucket in the $x$-$z$ plane during a successful episode in unseen material evaluation scenario. The green square denotes the proximity condition.}
\label{fig:TrajectorySandTestCondition}
\end{figure}

To gain deeper insight into the agent’s behavior, one successful episode is randomly selected for detailed analysis. 
The trajectories of the rock and the bucket in the $x$-$z$ plane are illustrated in Fig.~\ref{fig:TrajectorySandTestCondition}. Initially, the rock reaches the proximity of the goal, but it later drifts slightly outside the goal square due to motion along the $x$-axis. The bucket trajectory indicates that the agent adjusts its movement strategy to return the rock to the vicinity of the goal position.
\begin{figure}[htbp] 
\centering
\includegraphics[width=0.7\linewidth]{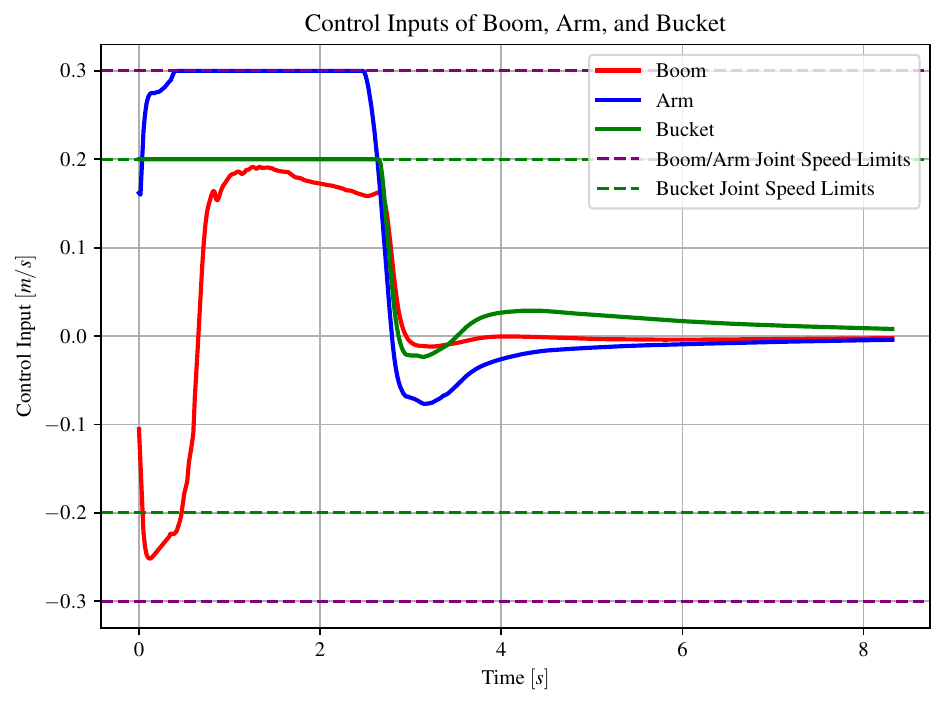}
\caption{Joint speed commands for the bucket, arm, and boom throughout a successful episode in unseen material evaluation scenario.}
\label{fig:ControlInputSandTestCondition}
\end{figure}
\begin{figure}[htbp] 
\centering
\includegraphics[width=0.7\linewidth]{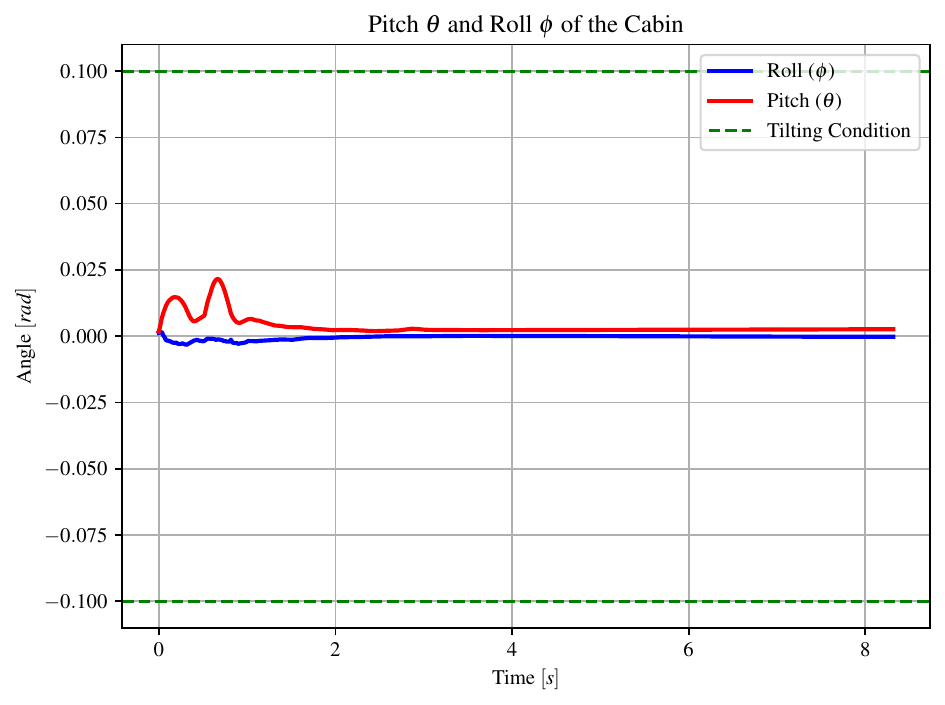}
\caption{Pitch $\theta$ and roll $\phi$ angles of the cabin during a successful episode in unseen material evaluation scenario. The valid range defined by the tilting condition $\mathbb{C}_{\text{tilting}}$ is indicated by green dashed lines.}
\label{fig:tiltingSandTestCondition}
\end{figure}
Figure~\ref{fig:ControlInputSandTestCondition} presents the joint speed commands for the bucket, arm, and boom during a successful episode in the unseen material evaluation scenario. Unlike the previous scenarios, around $t = 3~s$, the bucket joint moves in the negative direction before returning to a positive value. Such a pattern in the bucket speed command was not observed in two previous scenarios, suggesting that the agent adjusts its strategy to reposition the rock and maintain it within the goal proximity.
The cabin’s pitch $\theta$ and roll $\phi$ angles during a successful episode in the unseen material evaluation scenario are shown in Fig.~\ref{fig:tiltingSandTestCondition}.


\subsection{Comparison with Human Participants}\label{sec:HumanOperatorsCom}
In this evaluation, two human participants were invited to perform the rock capturing task under conditions identical to the training environment. The setup resembled a game-like interface, aligning with recent trends where simulators are commonly used to both train and assess the performance of human operators. Each participant was given the opportunity to practice in the AGX Dynamics\textsuperscript{\textregistered} simulator for $100$ trials. Following the practice phase, they completed $10$ evaluation trials each, resulting in a total of $20$ human-operated episodes. 
The first participant achieved a success rate of $0.8$, while the second participant reached $0.4$, yielding an overall success rate of $0.6$ for this scenario. These results highlight the difficulty of the task, even for human operators. It is important to note that comparing the RL agent with human operators is not straightforward. While the agent is explicitly trained to maximize the designed reward function, human operators pursue their own implicit objectives during task execution, which may not fully align with the agent’s reward structure. Despite this difference, humans are still relatively successful at completing the task, as reflected in their success rate.





\begin{figure}[htbp] 
\centering
\includegraphics[width=0.7\linewidth]{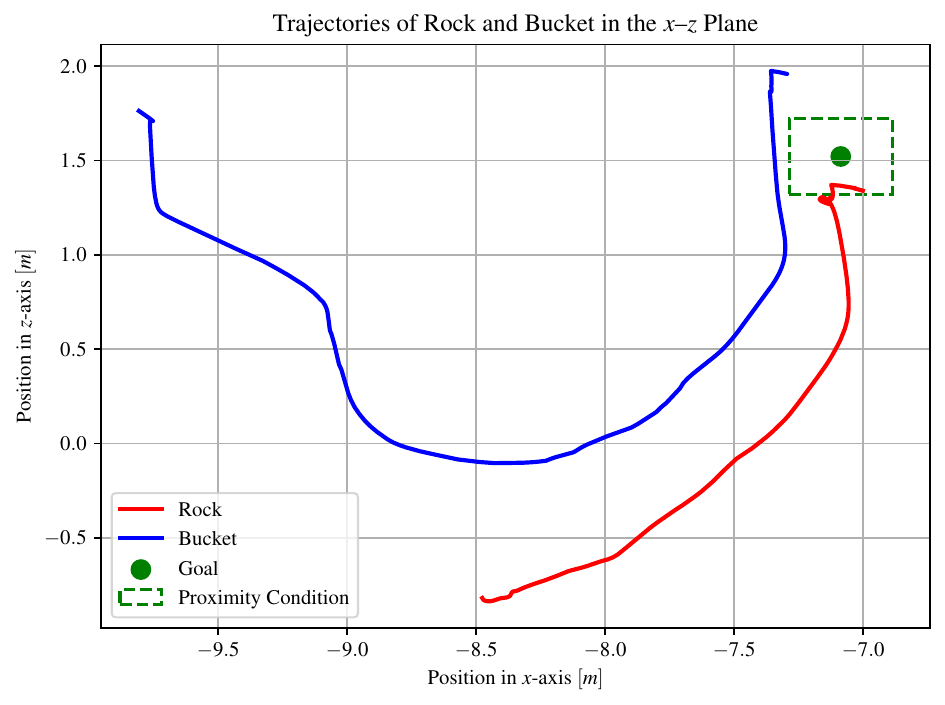}
\caption{Trajectories of the rock and bucket in the $x$-$z$ plane during a successful episode in human participant evaluation scenario. The green square denotes the proximity condition.}
\label{fig:TrajectoryHumanTestCondition}
\end{figure}

To better understand the behavior of human participants, one successful episode is randomly selected for detailed analysis. 
The trajectories of the rock and the bucket in $x$-$z$ plane are shown in Fig.~\ref{fig:TrajectoryHumanTestCondition}. It is important to note that unlike the learned agent, the human can see the goal location and its proximity area but does not have access to the exact position of the rock’s CoM. As a result, aligning the CoM precisely with the goal position is a challenging task. The human participant visually compares the entire rock and its geometry against the goal proximity area instead of focusing on a single reference point like the rock's CoM. Moreover, the operator interacts with the task through a 2D view of the simulator, which limits depth perception and spatial awareness. This restricted perspective further increases the difficulty of achieving precise alignment and can negatively affect overall performance compared to the agent.
\begin{figure}[htbp] 
\centering
\includegraphics[width=0.7\linewidth]{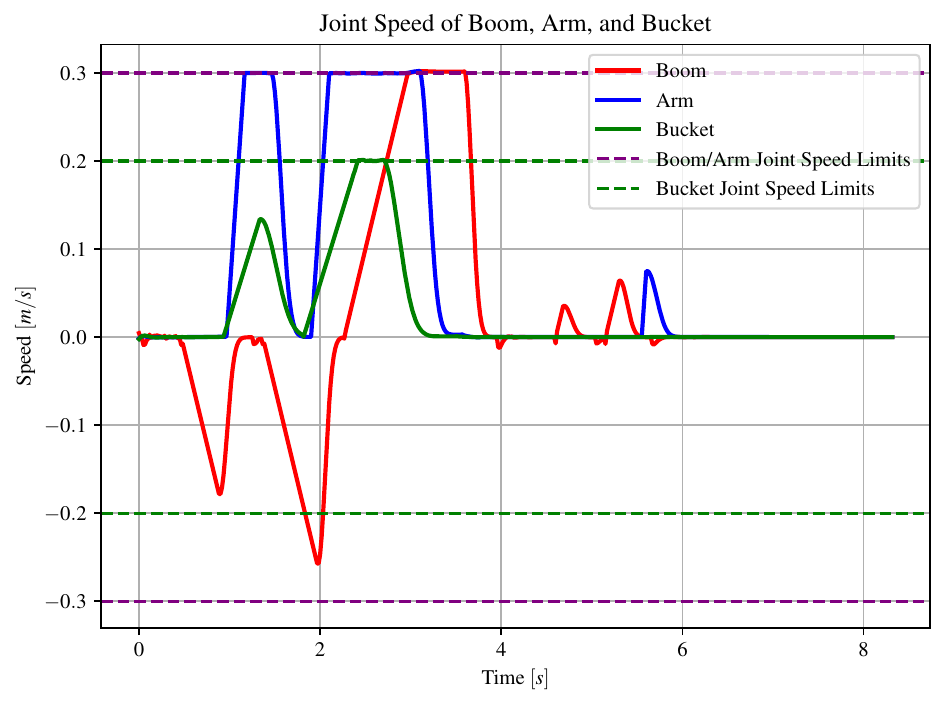}
\caption{Joint speed the bucket, arm, and boom throughout a successful episode in human participant evaluation scenario.}
\label{fig:ControlInputHumanTestCondition}
\end{figure}
\begin{figure}[htbp] 
\centering
\includegraphics[width=0.7\linewidth]{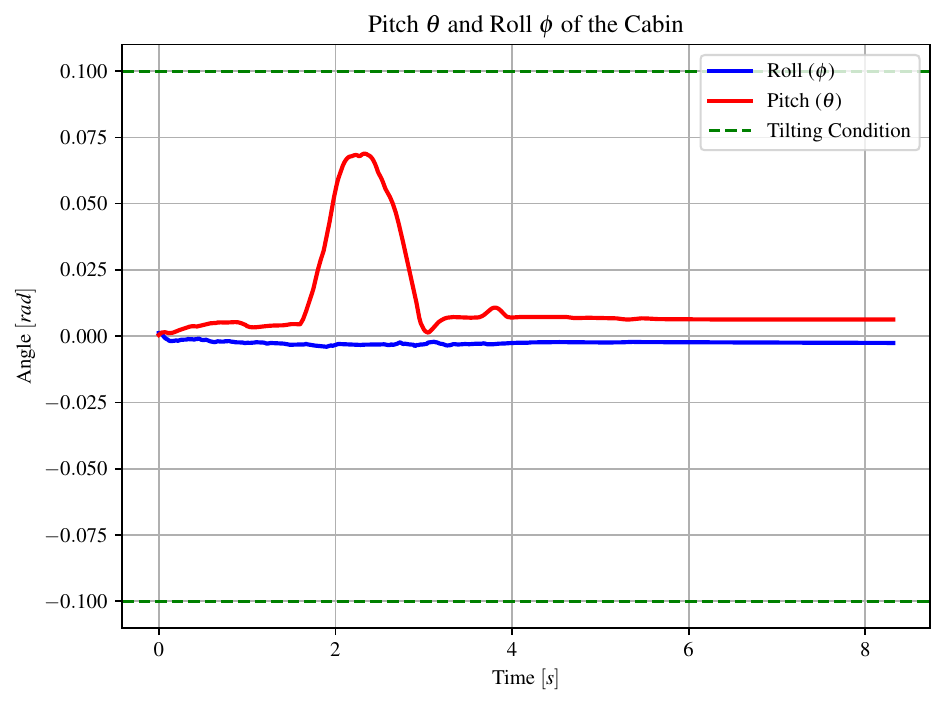}
\caption{Pitch $\theta$ and roll $\phi$ angles of the cabin during a successful episode in human participant evaluation scenario. The valid range defined by the tilting condition $\mathbb{C}_{\text{tilting}}$ is indicated by green dashed lines.}
\label{fig:tiltingHumanTestCondition}
\end{figure}
In this evaluation scenario, similar to a game-like simulator, human operators used a keyboard to send commands to control the speed of the bucket, arm, and boom joints. To enable comparison with other evaluation scenarios, the resulting joint speed derived from the keyboard inputs are visualized in Fig.~\ref{fig:ControlInputHumanTestCondition}. It is evident that the human control strategy differs significantly from that of the learned policy, exhibiting a bang–bang-like behavior in the joint space. This behavior can be attributed to two main factors: first, the limited sensitivity and resolution of the keyboard input device, and second, the lack of haptic feedback in the simulator, meaning that the operator does not feel the machine’s motion, vibrations, or contact forces. These factors make it challenging for human participants to apply smooth and continuous control actions, in contrast to the fine-grained control achievable by the RL policy. 
Finally, the cabin’s pitch $\theta$ and roll $\phi$ angles are shown in Fig.~\ref{fig:tiltingHumanTestCondition}. Compared to the learned policy across different evaluation scenarios, greater variation is observed in the pitch angle $\theta$, indicating less stable control by the human operator. However, the values remain within the acceptable tilting range.




\section{Discussion}\label{sec:Discussion}
This paper presented a RL-based controller for automatic rock capturing using an excavator. As demonstrated in the result section, the trained policy is capable of effectively capturing and moving the rock to the proximity of a desired goal location, even when the rock geometry or material properties differ from those used during training. As shown in Fig.~\ref{fig:SuccessRate}, the controller achieves a success rate of $0.8$, comparable to that of human participants. 

Despite these promising results, several limitations and areas for future improvement remain. The policy has only been evaluated in simulation. Deployment on a real machine is required to assess the sim-to-real transferability. While zero-shot transfer may be possible, fine-tuning on real-world data will likely be required. 
First, applying domain randomization to parameters related to the homogeneous material, rock properties, and contact interactions can reduce the sim-to-real gap. Additionally, reducing the particle size of the material could improve simulation realism and may inadvertently make the task easier, as smaller particles enable smoother interactions with the rock. However, this comes at the cost of increased computational complexity, which can slow down the training process.
Second, actuation delays can be explicitly modeled in the simulator to better reflect the latency present in real machine. 
Third, injecting noise into observations during training can help develop a more robust policy capable of handling real-world sensor inaccuracies. 
Forth, the simulation currently models the excavator’s joints as linear actuator with closed chain mechanisms, while real machines operate using hydraulic cylinders. Incorporating a more realistic hydraulic model would contribute to reducing the sim-to-real gap. 
Fifth, in the current setup, the rock’s CoM position is directly available from the simulator. However, in real-world applications, this information must be inferred using a vision-based system, which introduces uncertainty and noise. These discussions outline potential directions that will be further investigated in future work to enhance realism and facilitate real-world deployment.

From a computational perspective, the RL-based controller offers key advantages at deployment. Unlike online optimization-based methods, which often require solving computationally intensive problems at each step, the RL policy uses a simple neural network forward pass to generate control commands. This makes inference extremely fast and suitable for real-time applications. 

From a learning perspective, the current reward function does not penalize episode failure or timeouts. Introducing a negative reward in such cases could accelerate the learning process and encourage more efficient behaviors. Beyond success rate, energy consumption could also serve as a valuable secondary metric for evaluating the controller’s overall efficiency.
The reward function also does not penalize the amount of homogeneous soil scooped along with the rock. Including such a penalty could encourage more precise and efficient manipulation strategies; however, it may also make the learning process more challenging and slower.
Training efficiency could benefit from curriculum learning. Starting with relaxed goal conditions and gradually tightening them as training progresses would likely speed up convergence and improve performance.
Currently, the agent relies solely on the most recent observation. Incorporating a short history of past observations, either through observation stacking or by using recurrent neural networks such as Long Short-Term Memory (LSTM), could improve temporal reasoning and enhance performance. 

Safety is a critical consideration in the rock capturing task. Unsafe behaviors, such as aggressive or rapid bucket movements, can result in the rock being unintentionally thrown, posing serious risks to the machine and nearby personnel. Integrating safe RL techniques could enhance the stability of the system and reduce the likelihood of hazardous actions.
Finally, exploring alternative RL algorithms, such as Soft Actor-Critic (SAC), could potentially improve sample efficiency and performance. Additionally, integrating model-based RL approaches may further enhance learning speed and policy effectiveness by leveraging predictive models of the environment.

\section{Conclusion}\label{sec:Conclusion}

This paper presented a fully data-driven, RL-based control strategy for automating the rock capturing task using a standard excavator. Capturing and moving large rocks with a bucket is a highly skill-intensive task due to the unstructured environment and complex, dynamic interactions between the rock, granular material, and manipulator. Traditional control approaches struggle with these challenges because accurate analytical models are difficult to obtain and require extensive tuning. 
To address this, a model-free RL policy was trained entirely in the high-fidelity AGX Dynamics\textsuperscript{\textregistered} simulator using the PPO algorithm. The training employed domain randomization of rock geometry, density, mass, and initial configurations of the rock, bucket, and goal, ensuring robustness and generalization. The learned controller directly outputs joint speed commands for the boom, arm, and bucket, without relying on explicit knowledge of rock or material properties.  
Evaluation across multiple scenarios, including unseen rocks and varying material properties, demonstrates that the policy achieves a high success rate of $0.8$, comparable to human participants, while maintaining stable machine behavior and avoiding unsafe bucket motions. These results indicate that complex rock manipulation tasks can be effectively automated using standard excavator buckets through learning-based control, without requiring specialized end-effectors.  
Future work will focus on bridging the sim-to-real gap by incorporating hydraulic actuator modeling, actuation delays, vision-based rock tracking, and sensor noise injection. Safety-critical aspects, such as preventing aggressive bucket actions that could lead to hazardous outcomes, also warrant further exploration, potentially via safe RL techniques. To the best of our knowledge, this study is the first to propose and evaluate an RL-based controller for rock capturing with an excavator, highlighting the potential of learning-based methods for automating challenging earth-moving tasks in construction and mining operations.


\section*{Acknowledgement}\label{sec:acknow}
The work was funded in part by Horizon Europe Project XSCAVE under Grant 101189836, and in part by the Research Council of Finland through the PROFI 7 grant.



\section*{Declaration of Generative AI and AI-Assisted Technologies in the Writing Process}\label{sec:AITools}
During the preparation of this work the authors used ChatGPT in order to improve language clarity. After using this tool, the authors reviewed and edited the content as needed and take full responsibility for the content of the published article.

\section*{Declaration of Competing Interest}\label{sec:DecComInt}
The authors declare that they have no known competing financial interests or personal relationships that could have appeared to influence the work reported in this paper.

\appendix
\section{Episode Initialization Details}\label{appendix:A}

The rock is assigned a material, and its density is sampled from a normal distribution with a mean of $2000~kg/m^3$ and a standard deviation of $85~kg/m^3$. The rock is initialized along the $x$-axis of the base frame, with its $y$-coordinate set to $0$ and its $z$-coordinate elevated to $0.5~m$ above the ground to prevent initial collision with the terrain. The $x$-coordinate is sampled uniformly from the range $[-11.5, -8.0]$. 

\begin{figure}[htbp] 
    \centering
    \begin{subfigure}[b]{0.3\linewidth} 
        \centering
        \includegraphics[width=\linewidth]{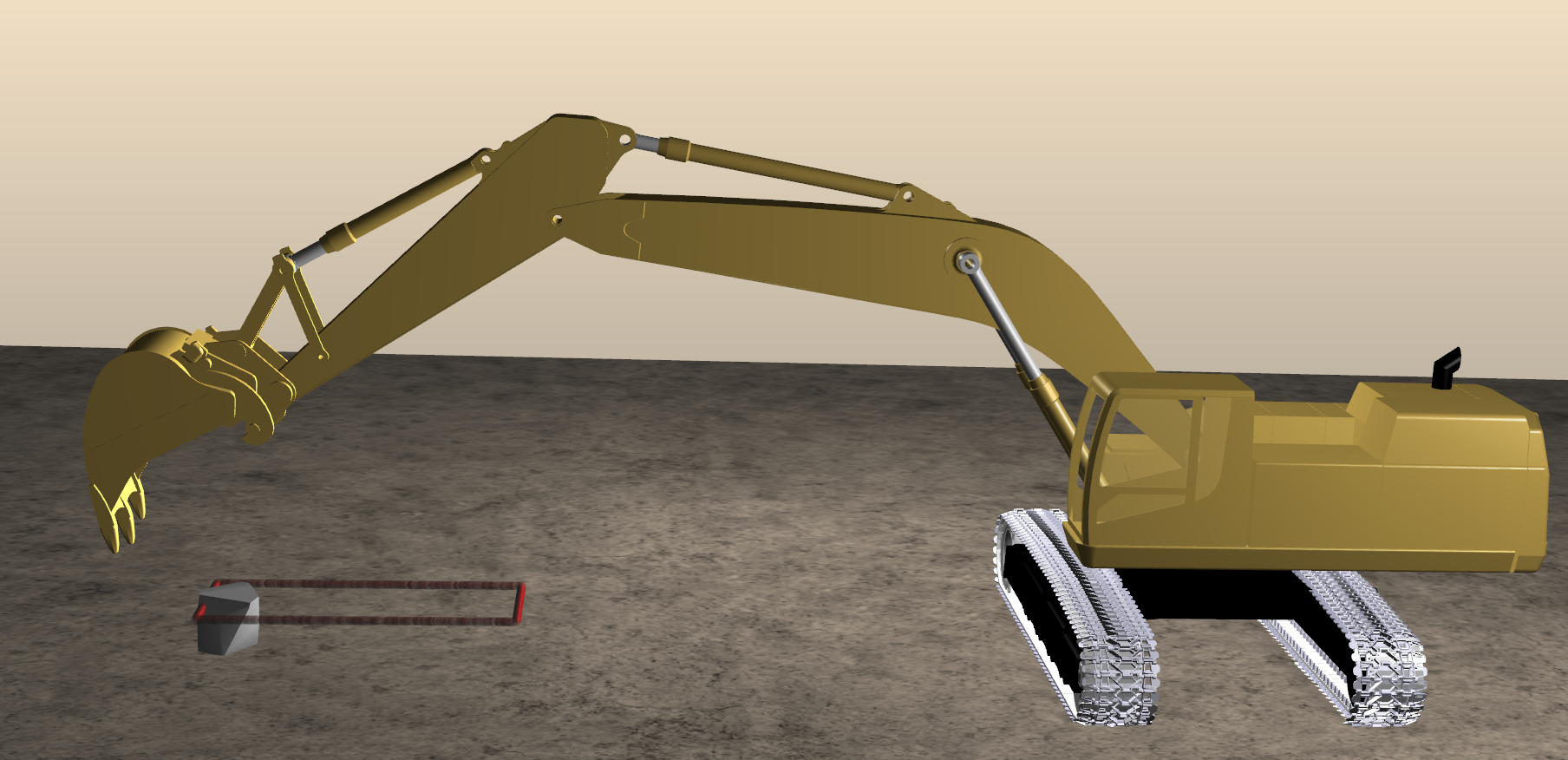}
        \caption{Rock near the far end of the effective workspace. }
        \label{fig:manipuConfig01}
    \end{subfigure}
    \hfill
    \begin{subfigure}[b]{0.3\linewidth} 
        \centering
        \includegraphics[width=\linewidth]{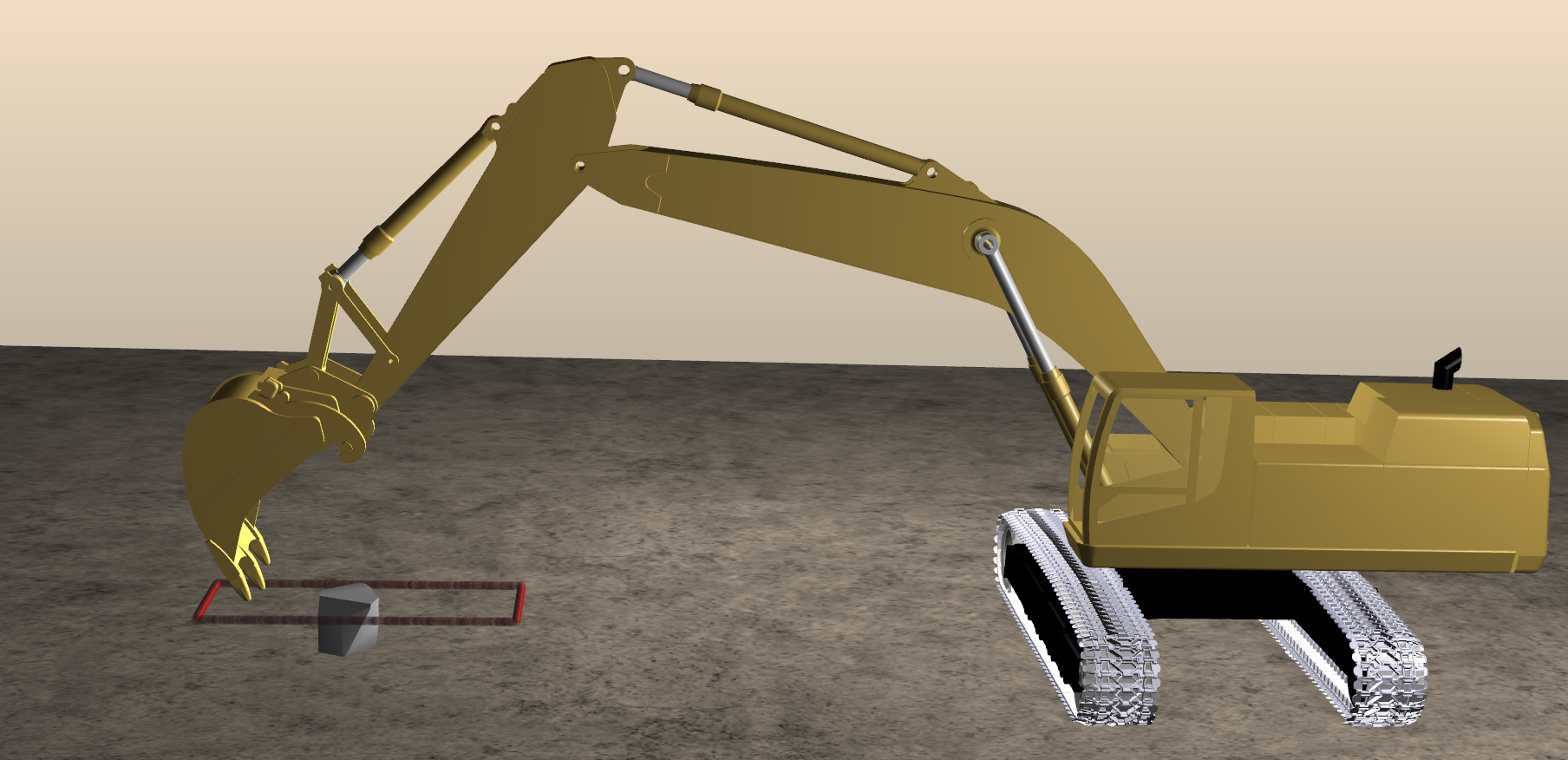} 
        \caption{Rock near the center of the effective workspace.}
        \label{fig:manipuConfig02}
    \end{subfigure}
    \hfill
    \begin{subfigure}[b]{0.3\linewidth} 
        \centering
        \includegraphics[width=\linewidth]{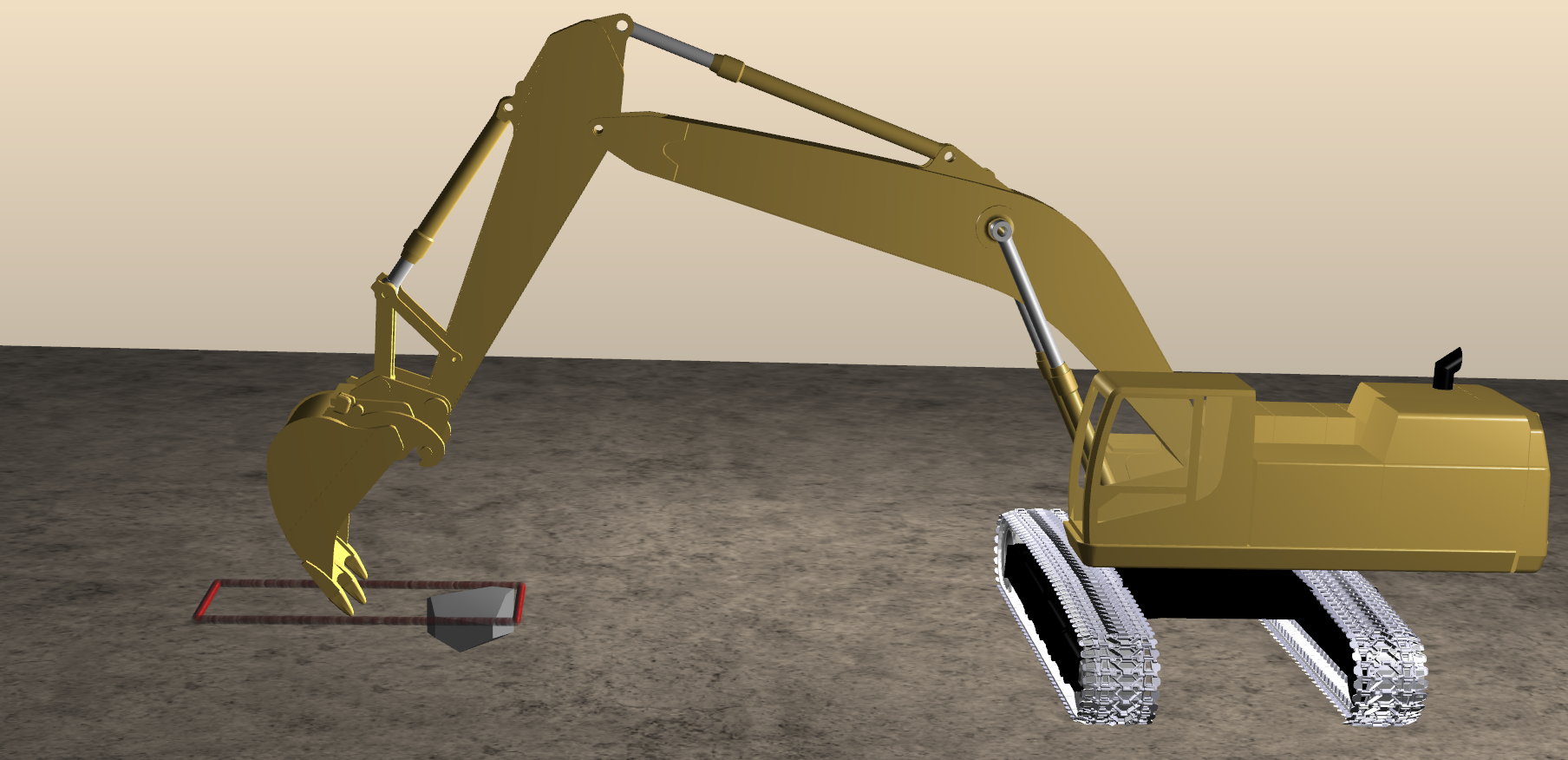} 
        \caption{Rock near the start of the effective workspace.}
        \label{fig:manipuConfig03}
    \end{subfigure}
    \caption{Schematic illustrations of three example initial configurations of the rock and the excavator’s bucket. The bucket is pre-positioned above and behind the rock before each episode begins. These configurations are determined based on the rock's initial position along the $x$-axis, as detailed in Table~\ref{tab:manipuConfig}.}
    \label{fig:manipuConfig}
\end{figure}

The initial configuration of the excavator’s bucket, arm, and boom is set such that the bucket starts behind and above the rock, positioned appropriately based on the rock’s position. Details of the initial manipulator configuration are provided in Table~\ref{tab:manipuConfig}.
\begin{table}[htbp]
    \centering
    \caption{Initial configurations of the excavator's boom, arm, and bucket based on the initial position of the rock in $x$-axis.}
    \label{tab:manipuConfig}
    \begin{tabular}{c c c c}
        \toprule
         Initial position of  & \multicolumn{3}{c}{Joint positions $[m]$} \\
         the rock in $x$-axis $[m]$ & Boom & Arm & Bucket \\
        \midrule
        $-8.0 \leq x_{\text{rock}}$          & +0.13 & +0.24 & -0.88 \\
        $-8.5 \leq x_{\text{rock}} < -8.0$   & +0.08 & +0.11 & -0.80 \\
        $-9.0 \leq x_{\text{rock}} < -8.5$   & +0.06 & -0.03 & -0.74 \\
        $-9.5 \leq x_{\text{rock}} < -9.0$   & +0.03 & -0.15 & -0.74 \\
        $-10.0 \leq x_{\text{rock}} < -9.5$   & -0.01 & -0.33 & -0.70 \\
        $-10.5 \leq x_{\text{rock}} < -10.0$   & -0.03 & -0.39 & -0.70 \\
        $-11.0 \leq x_{\text{rock}} < -10.5$  & -0.10 & -0.57 & -0.70 \\
        $-11.5 \leq x_{\text{rock}} < -11.0$ & -0.10 & -0.70 & -0.70 \\
        $x_{\text{rock}} < -11.5$            & -0.16 & -0.80 & -0.78 \\
        \bottomrule
    \end{tabular}
\end{table}
Note that moving the bucket to this initial position is not part of the RL agent's task; the agent begins from this pre-positioned state. A simple schematic of initial configurations of the rock and bucket is illustrated in Fig.~\ref{fig:manipuConfig}. 

It should be noted the terrain, which is made of dirt material with zero slope, is not subject to randomization. The dirt material represents a cohesive, moderately stiff soil commonly found in natural excavation environments. The key parameters used to define the dirt material are summarized in Table~\ref{tab:dirtMaterial}.
\begin{table}[htbp]
\centering
\caption{Key parameters of dirt material used in simulation.}
\begin{tabular}{lcc}
\toprule
Parameter & Value & Unit \\
\midrule
Cohesion (bulk) & $2100.0$ & Pa \\
Density & $1474.0$ & $kg/m^3$ \\
Maximum Density & $2000.0$ & $kg/m^3$\\
Young’s Modulus & $10^6$ & Pa \\
Internal Friction Angle & $0.70$ & $rad$ \\
Dilatancy Angle & $0.23$  & $rad$ \\
Swell Factor & $1.10$ & $-$ \\
Angle of Repose Compaction Rate & $24.0$ & $-$ \\
\bottomrule
\end{tabular}\label{tab:dirtMaterial}
\end{table}

\section{Sand Material Properties}\label{appendix:B}
The sand material used for the third evaluation scenario was defined to assess the generalization capability of the trained policy under different terrain conditions. The key physical parameters of the sand are summarized in Table~\ref{tab:sandMaterial}.
\begin{table}[ht]
\centering
\caption{Key parameters of sand material used in third evaluation scenario.}
\begin{tabular}{lcc}
\toprule
Parameter & Value & Unit \\
\midrule
Cohesion (bulk) & $0.0$ & Pa \\
Density & $1474.0$ & $kg/m^3$ \\
Maximum Density & $1800.0$ & $kg/m^3$\\
Young’s Modulus & $4.5\times10^6$ & Pa \\
Internal Friction Angle & $0.68$ & $rad$ \\
Dilatancy Angle & $0.16$  & $rad$ \\
Swell Factor & $1.0$ & $-$ \\
Angle of Repose Compaction Rate & $1.0$ & $-$ \\
\bottomrule
\end{tabular}\label{tab:sandMaterial}
\end{table}



\bibliographystyle{elsarticle-num} 

\bibliography{mybibfile}






\end{document}